\pdfoutput=1

\PassOptionsToPackage{table}{xcolor}

\documentclass[sigconf,nonacm]{acmart}

\usepackage{enumitem}
\usepackage{mathtools}
\usepackage{microtype}
\usepackage{subcaption}
\usepackage{booktabs}
\usepackage{multirow}
\usepackage{array}
\usepackage{makecell}
\usepackage{rotating}
\usepackage{afterpage} 
\usepackage{bm} 
\usepackage{etoolbox}
\usepackage{dblfloatfix}
\usepackage{placeins}
\usepackage{xspace}
\usepackage{algorithm}
\usepackage{algorithmic}
\usepackage{siunitx} 

\newcolumntype{B}{c}
\newcolumntype{T}{c}
\newcolumntype{O}{c}
\newcolumntype{b}{c}
\newcolumntype{t}{c}
\newcolumntype{o}{c}

\definecolor{MutedGray}{HTML}{6B7280}
\definecolor{BenchDatasetGray}{HTML}{F3F4F6}

\newcolumntype{V}{!{\color{black!55}\vrule width 0.45pt}}

\newcommand{\mstd}[2]{{#1{\scriptsize$_{\pm #2}$}}}

\newcommand{\best}[1]{{\bfseries\boldmath{}#1}}
\newcommand{\second}[1]{\underline{#1}}

\newcommand{\method}{\textsc{scHelix}\xspace}

\DeclareMathOperator{\Var}{Var}
\DeclareMathOperator{\ReLU}{ReLU}
\DeclareMathOperator{\softmax}{softmax}

\DeclareMathOperator{\LN}{LN}

\newcommand{\bbR}{\mathbb{R}}
\newcommand{\E}{\mathbb{E}}
\newcommand{\sg}{\operatorname{sg}}

\newcommand{\Gsel}{\mathcal{G}_{\mathrm{sel}}}
\newcommand{\Ginv}{\mathcal{G}_{\mathrm{inv}}}
\newcommand{\Gvar}{\mathcal{G}_{\mathrm{var}}}
\newcommand{\INV}{\mathrm{inv}}
\newcommand{\VAR}{\mathrm{var}}

\theoremstyle{definition}
\ifx\definition\undefined
  \newtheorem{definition}{Definition}[section]
\fi
\ifx\assumption\undefined
  \newtheorem{assumption}{Assumption}[section]
\fi
\theoremstyle{remark}
\ifx\remark\undefined
  \newtheorem{remark}{Remark}[section]
\fi

\usepackage{tikz}
\usetikzlibrary{arrows.meta}
\usepackage[most]{tcolorbox}

\definecolor{VariantRed}{HTML}{D84A4A}
\definecolor{AnchorBlue}{HTML}{3B82F6}
\definecolor{FusionPurple}{HTML}{7C3AED}

\newcommand{\duetAFigDir}{yxc_commit}
\newcommand{\duetAFig}[1]{\duetAFigDir/#1}

\newcommand{\duetAMaybeInclude}[3][]{%
  \IfFileExists{\duetAFig{#2}}%
    {\includegraphics[#1]{\duetAFig{#2}}}%
    {\includegraphics[#1]{\duetAFig{#3}}}%
}

\tcbset{
  duetA_panel/.style={
    enhanced, colback=white, colframe=black!12, arc=1.6mm, boxrule=0.45pt,
    left=1.2mm,right=1.2mm,top=1.0mm,bottom=1.0mm,
  },
  duetA_plot/.style={
    enhanced, colback=white, colframe=black!10, arc=1.0mm, boxrule=0.35pt,
    left=0.7mm,right=0.7mm,top=0.6mm,bottom=0.6mm,
  },
  duetA_tag/.style={
    on line, colback=white, colframe=black!22, arc=0.9mm, boxrule=0.3pt,
    left=0.9mm,right=0.9mm,top=0.15mm,bottom=0.15mm,
  }
}

\usepackage[capitalize,noabbrev]{cleveref}

\begin{document}

\title{\method: Asymmetric Dual-Stream Integration via Explicit Gene-Level Disentanglement}

\author{Xichen Yan}
\authornote{Equal contribution.}
\affiliation{%
  \institution{Jinan University}
  \city{Guangzhou}
  \country{China}
}

\author{Zelin Zang}
\authornotemark[1]
\authornote{Corresponding author: \texttt{zangzelin@westlake.edu.cn}}
\affiliation{%
  \institution{Westlake University}
  \city{Hangzhou}
  \country{China}
}

\author{Changxi Chi}
\affiliation{%
  \institution{Westlake University}
  \city{Hangzhou}
  \country{China}
}

\author{Jingbo Zhou}
\affiliation{%
  \institution{Westlake University}
  \city{Hangzhou}
  \country{China}
}

\author{Chang Yu}
\affiliation{%
  \institution{Westlake University}
  \city{Hangzhou}
  \country{China}
}

\author{Jinlin Wu}
\affiliation{%
  \institution{HKISI}
  \city{Hong Kong}
  \country{China}
}

\author{Shenghui Cheng}
\affiliation{%
  \institution{Jinan University}
  \city{Guangzhou}
  \country{China}
}

\author{Fuji Yang}
\affiliation{%
  \institution{TIAS}
  \city{Hangzhou}
  \country{China}
}

\author{Jiebo Luo}
\affiliation{%
  \institution{HKISI}
  \city{Hong Kong}
  \country{China}
}

\author{Zhen Lei}
\affiliation{%
  \institution{HKISI}
  \city{Hong Kong}
  \country{China}
}

\author{Stan Z. Li}
\affiliation{%
  \institution{Westlake University}
  \city{Hangzhou}
  \country{China}
}

\renewcommand{\shortauthors}{Yan, Zang et al.}

\begin{abstract}
    A critical challenge in single-cell RNA sequencing (scRNA-seq) integration is resolving the tension between eliminating batch effects and maintaining biological fidelity. While recent evidence indicates that batch effects manifest heterogeneously across genes, most existing methods process the transcriptome uniformly, frequently resulting in over-correction and loss of subtle biological signals. To address this, we present \method, a dataset-adaptive framework that fundamentally changes how features are processed by explicitly partitioning genes into domain-invariant Anchors and domain-sensitive Variants at the input level. \method utilizes a dual-stream sparse diffusion encoder equipped with stop-gradient graph caching to efficiently learn multi-scale structural representations. The core of our approach is a novel asymmetric Align-Refine-Fuse protocol: the unstable Variant stream is first aligned to the robust topology of the Anchor stream, followed by a conservative refinement phase where the Anchor stream absorbs denoised details via bounded residual gating. This ``divide-and-conquer'' architecture prevents shortcut learning and ensures robust batch removal without compromising the integrity of biological clusters. Extensive benchmarking demonstrates that \method outperforms state-of-the-art methods.The source code is available at \url{https://anonymous.4open.science/status/scHelix-E175}
\end{abstract}


\maketitle

\section{Introduction}
\label{sec:intro}

\begin{figure}[t]
    \centering
    \includegraphics[width=0.99\columnwidth]{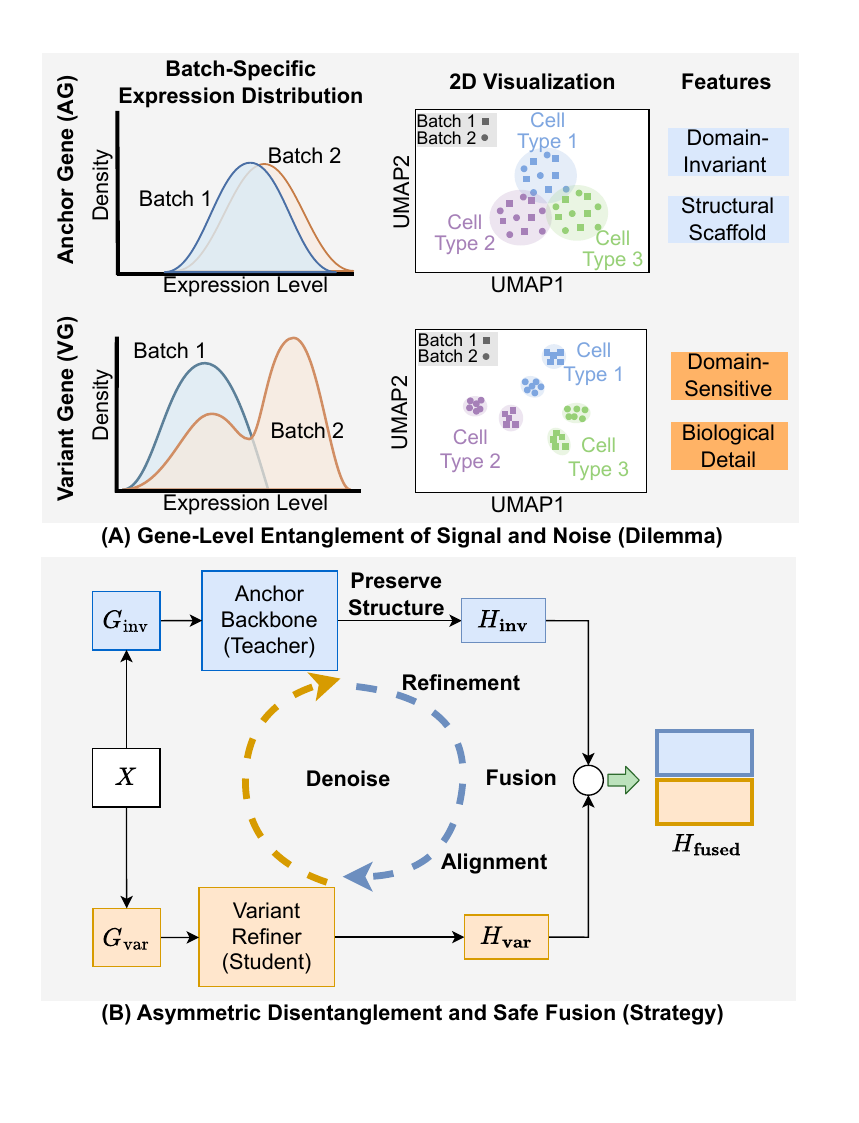}
    \vspace{-3.5em}
    \caption{\textbf{Motivation and Conceptual framework of \method.} \textbf{(A) The Dilemma}: Batch effects are gene-specific. \textbf{Anchor Genes (AG)} serve as a stable scaffold, while \textbf{Variant Genes (VG)} are prone to domain noise. \textbf{(B) The Strategy}: \method partitions genes into invariant ($G_{\text{inv}}$) and variant ($G_{\text{var}}$) streams. The Anchor Backbone (Teacher) guides the Variant Refiner (Student) to align and denoise, preserving topology in the fused representation ($H_{\text{fused}}$).}
    \label{fig:motivation}
    \Description{Conceptual diagram showing the scHelix motivation: panel A illustrates gene-specific batch effects with Anchor Genes being stable and Variant Genes being noisy across domains; panel B shows the dual-stream strategy where an invariant Anchor backbone guides a Variant Refiner to produce a fused representation.}
\end{figure}

Single-cell RNA sequencing (scRNA-seq) integration is the cornerstone of constructing unified cell atlases, unlocking the potential to map dynamic trajectories across diverse populations~\cite{trapnell2015defining, luecken2022benchmarking}. However, it faces a fundamental trade-off: removing technical batch effects often comes at the cost of erasing subtle biological heterogeneity~\cite{leek2010tackling, haghverdi2018batch}. Aggressive correction methods tend to over-smooth data, blurring fine-grained cell states, while conservative approaches fail to align shared populations across domains~\cite{luecken2022benchmarking}.

This tension persists because most existing methods treat the transcriptome uniformly, processing the entire gene space (e.g., HVGs) as a monolithic feature vector~\cite{lopez2018deep, korsunsky2019harmony, stuart2019comprehensive}. This ignores a critical reality: batch effects are highly \emph{gene-specific}~\cite{zhou2025gte}.
As illustrated in \Cref{fig:motivation}, genes governing core cellular programs—such as cell-cycle regulators (e.g., \textit{PCNA})—often maintain consistent distributions across platforms, providing a robust scaffold for cell identity. Conversely, context-sensitive genes—such as the interferon-stimulated gene \textit{ISG15}—carry genuine biological signals but are simultaneously confounded by technical artifacts.
By forcing these conflicting signals into a single latent space, existing models often settle for a brittle compromise: either residual batch effects or loss of biological fidelity~\cite{luecken2022benchmarking}.

To address these challenges, we argue that integration methods must move beyond implicit latent space regularization to \emph{explicit feature-level disentanglement}~\cite{lotfollahi2019scgen, piran2024disentanglement}. Specifically, we answer three core questions:
(1) \textbf{Unsupervised Anchor Discovery}: How can we identify a feature scaffold that is biologically discriminative yet technically stable, without relying on external annotations?
(2) \textbf{Structural Disentanglement}: How do we prevent domain-sensitive noise from dominating the representation learning process (i.e., avoiding shortcut learning)?
(3) \textbf{Safe Interaction}: How can we recover fine-grained details from noisy features without reintroducing batch artifacts into the stable topology?

We propose \textbf{\method}, a "divide-and-conquer" framework that explicitly disentangles features at the input level. 
Instead of a uniform input, \method partitions genes into domain-invariant \textbf{Anchors} and domain-sensitive \textbf{Variants} using a dataset-adaptive discriminability-sensitivity criterion (\Cref{sec:disentangle}). This acts as a \emph{self-supervised feature selector}, creating a hard information barrier: the Anchor stream serves as a stable "Teacher" to preserve topology, while the Variant stream acts as a "Student" tasked with denoising. 

Unlike MLP-based methods that treat cells as independent samples~\cite{lopez2018deep, xiong2022scalex}, \method employs a \textbf{dual-stream sparse diffusion encoder} to capture the non-linear manifold structure of the data~\cite{wang2021scgnn, klicpera2019appnp}.
Crucially, we introduce a novel asymmetric \textbf{Align-Refine-Fuse} protocol. First, the unstable Variant stream aligns to the robust Anchor topology. Second, the Anchor stream conservatively absorbs denoised details via bounded residual gating. Finally, the streams are fused via a hyper-network. This serial, controlled interaction ensures robust batch removal without compromising the integrity of biological clusters.

Our contributions are:
\begin{itemize}[leftmargin=*]
    \item \textbf{Explicit Disentanglement}: A dataset-adaptive partition strategy that separates stable anchors from noisy variants, providing an interpretable biological scaffold.
    \item \textbf{Asymmetric Graph Architecture}: A dual-stream design that leverages graph diffusion to learn multi-scale structural representations while strictly isolating noise sources.
    \item \textbf{Align-Refine-Fuse Protocol}: A conservative interaction mechanism that balances topology preservation with detail recovery, preventing anchor contamination.
    \item \textbf{SOTA Performance}: Extensive benchmarks demonstrate that \method achieves a superior trade-off between batch mixing and biological conservation compared to state-of-the-art baselines.
\end{itemize}

\section{Related Work}
\label{app:related}

\paragraph{Batch correction and feature heterogeneity.}
Single-cell integration seeks to remove technical batch effects while preserving biological
structure.
Classical methods align datasets via cross-domain neighbor or anchor matching—e.g., MNN,
Seurat, and Harmony~\cite{haghverdi2018batch,korsunsky2019fast,stuart2019comprehensive}—while
deep generative models such as scVI scale integration by learning batch-conditioned latent
spaces~\cite{lopez2018deep}.
However, most approaches treat the transcriptome as a uniform feature vector and apply
global alignment pressure, implicitly assuming batch effects are globally distributed.
Recent gene-level analyses challenge this assumption: the Group Technical Effect (GTE)
metric reveals that batch shifts are strongly gene-dependent and often dominated by a small
subset of highly batch-sensitive genes~\cite{zhou2025quantifying}, motivating feature-aware
correction strategies.

\paragraph{Disentanglement, dual-stream architectures, and self-supervised alignment.}
Disentangled representation learning—via latent regularizers
(scGen, biolord)~\cite{lotfollahi2019scgen,piran2024disentanglement} or causal-inspired
factorized blocks
(CRADLE-VAE, FCR, SCBD)~\cite{baek2024cradle,makino2025supervised,mao2024learning}—aims
to separate biological from nuisance variation, yet implicit enforcement can be difficult
to audit and may still permit leakage.
Dual encoders (e.g., BioBatchNet)~\cite{liu2025biobatchnet} provide explicit architectural
separation, asymmetric self-supervised objectives
(BYOL/SimSiam)~\cite{chen2021exploring,grill2020bootstrap} offer stable teacher targets,
and graph-based encoders (e.g., scGNN)~\cite{wang2021scgnn} capture single-cell topology.
Our method builds on these ideas but makes disentanglement \emph{explicit at the input
level} via Anchor/Variant gene partitioning, coupled with a conservative
Align--Refine--Fuse protocol that transfers denoised details while protecting the
invariant scaffold.

\section{Method}
\label{sec:method}
\begin{figure*}[t]
    \centering
  
    \includegraphics[width=1.07\textwidth]{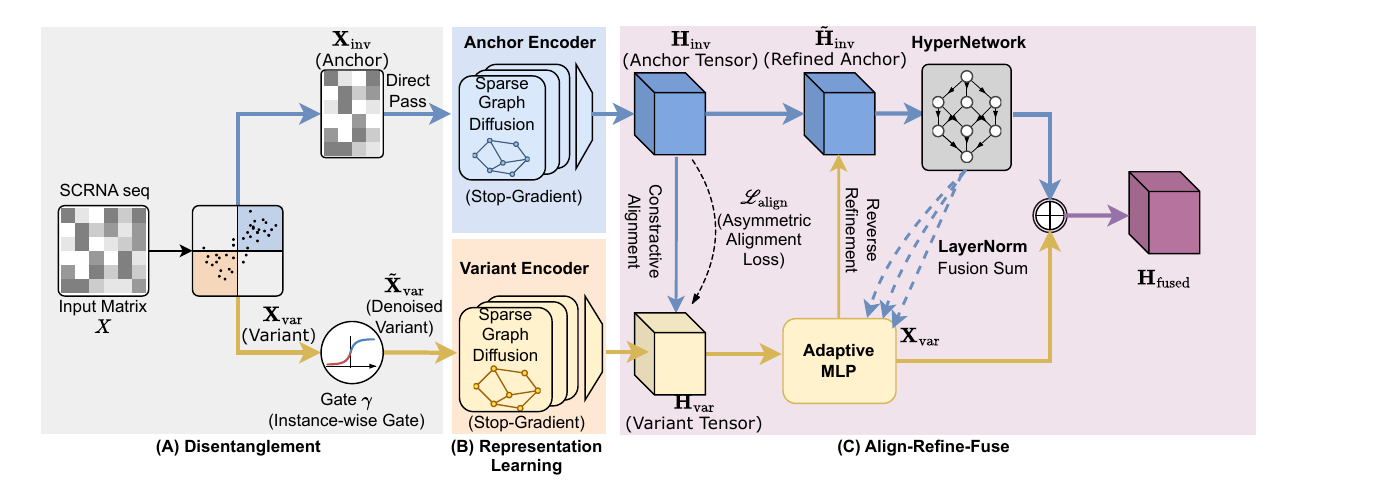}
    \vspace{-3.5em}
    \caption{\textbf{scHelix overview.}
    \textbf{(A) Disentanglement:} input-level feature split 
    (Anchors vs.\ Variants) via a discriminability--sensitivity 
    quadrant, with an instance-wise gate applied \emph{only} 
    to Variants (Section~\ref{sec:disentangle}).
    \textbf{(B) Representation Learning:} dual-stream 
    multi-scale \emph{graph diffusion} encoders on learnable 
    sparse feature graphs, with stop-gradient graph caching 
    for efficiency and stability (Section~\ref{sec:graph_encoder}).
    \textbf{(C) Align-Refine-Fuse:} serial 
    Align\(\rightarrow\)Refine interaction establishes a stable 
    Anchor topology before bounded residual transfer, 
    followed by HyperFusion and a Teacher-guided 
    warm-up\(\rightarrow\)fusion schedule 
    (Section~\ref{sec:serial_fusion}).
    All protective mechanisms (hard barrier, stop-gradient, 
    bounded residual) prevent anchor contamination while 
    enabling controlled information flow.}
    \label{fig:architecture}
    \Description{Architecture overview of scHelix showing three stages: left panel depicts input-level feature split into Anchors and Variants with an instance-wise gate; middle panel shows dual-stream graph diffusion encoders on sparse feature graphs; right panel illustrates the serial Align-Refine interaction followed by HyperFusion.}
\end{figure*}

\subsection{Problem Setup}
\label{sec:setup}

We formulate multi-domain scRNA-seq integration as representation learning under domain shift \citep{gong2016domain}. We adopt standard terminology: an \emph{instance} is a cell, a \emph{feature} is a gene, and a \emph{domain} is a batch/technology channel. Given an expression matrix \(\mathbf{X}\in\bbR^{N\times G}\) and domain labels \(b_i\in\mathcal{B}\) for each instance \(i\in\{1,\dots,N\}\), our goal is to learn an embedding \(\mathbf{h}_i\in\bbR^{d}\) that preserves biological neighborhood structure while suppressing domain-specific variation. We follow a standard Scanpy-style preprocessing pipeline~\citep{wolf2018scanpy} (Appendix~\ref{app:preproc}). Throughout, \(\sg(\cdot)\) denotes stop-gradient and \(\LN(\cdot)\) denotes LayerNorm; let \(\varepsilon>0\) be a small constant, \(\|\cdot\|_2\) the \(\ell_2\) norm, \(\operatorname{norm}(\mathbf{u})\coloneqq \mathbf{u}/(\|\mathbf{u}\|_2+\varepsilon)\), and \([x]_0^1\coloneqq \min\{1,\max\{0,x\}\}\).

\subsection{Feature Disentanglement \& Variant Gating}
\label{sec:disentangle}

\textbf{Feature partition with a hard information barrier 
(\Cref{fig:architecture}(A)).}
To operationalize our ``divide-and-conquer'' strategy, we first split the informative gene pool \(\Gsel\) (e.g., HVGs) into two disjoint sets: \(\Gsel=\Gvar\cup\Ginv\). Specifically, \(\Ginv\) denotes \emph{domain-invariant Anchors} (the stable \textbf{Teacher} stream), and \(\Gvar\) captures \emph{domain-sensitive Variants} (the noisy \textbf{Student} stream). We reorder genes so that \(\Gvar\) occupies the first \(G_{\VAR}=|\Gvar|\) dimensions; accordingly, each cell input \(\mathbf{x}_i\) is physically split as \(\mathbf{x}_i=[\mathbf{x}^{\VAR}_i\,\|\,\mathbf{x}^{\INV}_i]\). Crucially, this creates a \emph{hard information barrier}: each stream receives only its specific feature subset, preventing the Variant stream from shortcut-copying stable features and forcing it to learn denoising.

\textbf{Anchor discovery via a discriminability--sensitivity quadrant.}
We identify these sets using a dataset-adaptive quadrant. For each gene \(g\in\Gsel\), we compute (i) a \emph{domain sensitivity} score \(s_g^{\mathrm{dom}}\) (Eq.~\ref{eq:domain_sens}) and (ii) an unsupervised \emph{structure separability} score \(s_g^{\mathrm{str}}\) (Eq.~\ref{eq:struct_disc}).
For domain sensitivity, let \(\mu_g^{(b)}\) be the mean expression in domain \(b\), and \(\mu_g^{(\mathrm{all})},\sigma_g^{(\mathrm{all})}\) be the global statistics:
\begin{equation}
    \label{eq:domain_sens}
    s_g^{\mathrm{dom}}
    \coloneqq
    \frac{1}{|\mathcal{B}|}
    \sum_{b\in\mathcal{B}}
    \frac{\big|\mu_g^{(b)}-\mu_g^{(\mathrm{all})}\big|}{\sigma_g^{(\mathrm{all})}+\varepsilon}.
\end{equation}
We z-score \(\{s_g^{\mathrm{dom}}\}\) to obtain \(z_g^{\mathrm{dom}}\). For structure separability, we generate pseudo-clusters \(c(i)\) via PCA\(\rightarrow\)kNN\(\rightarrow\)Leiden \citep{traag2019from} on \(\mathbf{X}_{[:,\Gsel]}\) and define:
\begin{equation}
    \label{eq:struct_disc}
    s_g^{\mathrm{str}}
    \coloneqq
    \frac{\Var_{\mathrm{between}}(g)}{\Var_{\mathrm{within}}(g)+\varepsilon}.
\end{equation}
After log-transforming and z-scoring to obtain \(z_g^{\mathrm{str}}\), we apply the quadrant rule:
\begin{align}
    \label{eq:quadrant_rule}
    \Ginv & \coloneqq
    \bigl\{g\in\Gsel: z_g^{\mathrm{dom}}\le\tau_{\mathrm{dom}},\, z_g^{\mathrm{str}}\ge\tau_{\mathrm{str}}\bigr\}, \\
    \Gvar & \coloneqq \Gsel\setminus\Ginv. \notag
\end{align}
Unless stated otherwise, we use \((\tau_{\mathrm{dom}},\tau_{\mathrm{str}})=(0,0)\).
Because both scores are z-scored within each dataset, this fixed threshold in standardized space corresponds to a \emph{dataset-specific} mean-bisection in raw-score space: a gene qualifies as an Anchor if and only if its domain sensitivity falls below the dataset mean \emph{and} its structure separability exceeds the dataset mean.
This mean-bisection rule provides a statistically natural, parameter-free default that automatically adapts to each dataset's distributional characteristics; we provide a formal stability analysis (boundary-band argument) in Appendix~\ref{app:threshold}.
This ensures that genes routed to the Anchor stream are both technically stable and biologically informative. Genes that are biologically relevant but batch-affected are routed to the Variant stream to be rectified, rather than discarded.

\textbf{Variant-only instance-wise domain gate.}
To handle cell-state-dependent batch effects without altering the stable scaffold, we apply an instance-wise gate \emph{only} to the Variant stream. We estimate a local domain sensitivity \(\gamma_i(g)\in[0,1]\) by mixing domain sensitivity scores from multi-resolution clusterings (details in Appendix~\ref{app:gate_defs}). The gated Student input is defined using an indicator function \(\mathbb{I}_{\{\cdot\}}\):
\begin{equation}
    \label{eq:attenuate}
    \tilde{x}_{i,g} \coloneqq x_{i,g} \left( 1 - \mathbb{I}_{\{g \in \Gvar\}} \cdot \lambda \cdot \gamma_i(g) \right).
\end{equation}
where \(\lambda\in[0,1]\) controls suppression strength. This mechanism precisely attenuates technical noise in the Variant stream while preserving the integrity of the Anchor stream.

\subsection{Representation Learning via Dual-Stream Sparse Graph Diffusion}
\label{sec:graph_encoder}

\begin{table*}[t]
  \centering
  \caption{\textbf{Main benchmark results across three datasets (mean$\pm$std).}
    Bio-conservation includes structure (ASW$_{ct}$, GC) and cluster agreement (ARI, NMI).
    We report $\mathrm{BioMean}=\tfrac{1}{4}(\mathrm{ARI}_{\mathrm{best}}+\mathrm{NMI}_{\mathrm{best}}+\mathrm{ASW}_{\mathrm{ct}}+\mathrm{GC})$ and $\mathrm{Overall}=0.4\,\mathrm{ASW}_{\mathrm{batch}}+0.6\,\mathrm{BioMean}$,
    following the scIB benchmark protocol~\cite{luecken2022benchmarking}. %
    All metrics are scaled to [0,\,1]; higher is better.
    \textbf{Bold}: best; \underline{underline}: second best.
    The last column (\textbf{Gain}) shows the relative improvement of \method over the second-best method in Overall score.}
    \vspace{-1.5em}
    \label{tab:main_benchmark_compact_yearsorted}
  \renewcommand{\mstd}[2]{#1{\scriptsize\,\textcolor{gray!60}{$\pm$#2}}}
  \newcommand{\score}[1]{#1} %
  \newcolumntype{g}{c}
  \newcolumntype{h}{c} %
  \resizebox{\linewidth}{!}{%
    \setlength{\tabcolsep}{6pt}
    \begin{tabular}{@{} l c c c c c c c c g h @{}}
      \toprule
      & 
      \multicolumn{2}{c}{\textbf{Info}} & 
      \multicolumn{5}{c}{\textbf{Bio-conservation} ($\uparrow$)} & 
      & 
      \multicolumn{1}{c}{} & 
      \multicolumn{1}{c}{} \\
      \cmidrule(lr){2-3} \cmidrule(lr){4-8}
      & \multirow{2}{*}{\textbf{Year}} & \multirow{2}{*}{\textbf{Type}} & 
      \multicolumn{2}{c}{\textbf{Structure}} & 
      \multicolumn{2}{c}{\textbf{Cluster}} & 
      \textbf{Avg.} & & 
      \multicolumn{1}{c}{} & 
      \multicolumn{1}{c}{} \\
      \cmidrule(lr){4-5} \cmidrule(lr){6-7} \cmidrule(lr){8-8}
      \multirow{-3}{*}{\textbf{Method}} & & & ASW$_{ct}$ & GC & ARI & NMI & BioMean & 
      \multirow{-3}{*}{\textbf{ASW$_{batch}$} ($\uparrow$)} & 
      \multicolumn{1}{c}{\multirow{-3}{*}{\textbf{Overall} ($\uparrow$)}} & 
      \multicolumn{1}{c}{\multirow{-3}{*}{\textbf{Gain} ($\uparrow$)}} \\
      \midrule
      \rowcolor{gray!10}
      \multicolumn{11}{c}{\textbf{Human Pancreas}}                                                                                            \\
      scVI             & 2018                                                       & Deep Learning
                       & \mstd{0.567}{0.003}                                        & \mstd{0.856}{0.012}
                       & \mstd{0.940}{0.002}                                        & \mstd{0.914}{0.002}
                       & 0.819
                       & \mstd{0.898}{0.003}                                        & \mstd{0.851}{0.003}                                     & - \\
      Harmony          & 2019                                                       & Traditional
                       & \mstd{0.649}{0.000}                                        & \second{\mstd{0.919}{0.001}}
                       & \second{\mstd{0.946}{0.000}}                               & \mstd{0.912}{0.001}
                       & 0.857
                       & \mstd{0.882}{0.000}                                        & \mstd{0.867}{0.000}                                     & - \\
      INSCT            & 2021                                                       & Deep Learning
                       & \mstd{0.663}{0.002}                                        & \mstd{0.900}{0.032}
                       & \mstd{0.931}{0.005}                                        & \mstd{0.885}{0.014}
                       & 0.845
                       & \mstd{0.796}{0.006}                                        & \mstd{0.825}{0.007}                                     & - \\
      SCALEX           & 2022                                                       & Deep Learning
                       & \mstd{0.631}{0.003}                                        & \mstd{0.901}{0.009}
                       & \mstd{0.910}{0.000}                                        & \mstd{0.901}{0.009}
                       & 0.836
                       & \mstd{0.877}{0.004}                                        & \mstd{0.857}{0.000}                                     & - \\
      sysVI            & 2024                                                       & Deep Learning
                       & \second{\mstd{0.694}{0.012}}                               & \best{\mstd{0.962}{0.004}}
                       & \mstd{0.946}{0.003}                                        & \mstd{0.907}{0.005}
                       & \best{0.877}
                       & \second{\mstd{0.902}{0.004}}                               & \second{\mstd{0.887}{0.001}}                            & - \\
      SCEMENT          & 2025                                                       & Traditional
                       & \mstd{0.617}{0.000}                                        & \mstd{0.912}{0.002}
                       & \mstd{0.945}{0.001}                                        & \second{\mstd{0.915}{0.001}}
                       & 0.847
                       & \mstd{0.890}{0.000}                                        & \mstd{0.864}{0.000}                                     & - \\
      scCobra          & 2025                                                       & Deep Learning
                       & \mstd{0.606}{0.003}                                        & \mstd{0.903}{0.013}
                       & \mstd{0.945}{0.001}                                        & \mstd{0.909}{0.002}
                       & 0.841
                       & \mstd{0.873}{0.005}                                        & \mstd{0.854}{0.000}                                     & - \\
      \rowcolor{blue!3}
      \textbf{\method} & \textbf{Ours}                                              & Deep Learning
                       & \best{\mstd{0.710}{0.005}}                                 & \mstd{0.913}{0.001}
                       & \best{\mstd{0.950}{0.001}}                                 & \best{\mstd{0.922}{0.002}}
                       & \second{0.874}
                       & \best{\mstd{0.927}{0.003}}                                 & \best{\mstd{0.895}{0.002}}                              & \textbf{+0.90\%} \\

      \addlinespace[0.4em]
      \rowcolor{gray!10}
      \multicolumn{11}{c}{\textbf{Failing Human Heart}}                                                                                       \\
      scVI             & 2018                                                       & Deep Learning
                       & \mstd{0.524}{0.001}                                        & \mstd{0.818}{0.001}
                       & \mstd{0.964}{0.005}                                        & \second{\mstd{0.953}{0.006}}
                       & 0.815
                       & \mstd{0.873}{0.005}                                        & \mstd{0.838}{0.002}                                     & - \\
      Harmony          & 2019                                                       & Traditional
                       & \mstd{0.720}{0.000}                                        & \mstd{0.856}{0.005}
                       & \second{\mstd{0.965}{0.000}}                               & \mstd{0.945}{0.003}
                       & 0.872
                       & \mstd{0.861}{0.000}                                        & \mstd{0.867}{0.001}                                     & - \\
      INSCT            & 2021                                                       & Deep Learning
                       & \mstd{0.602}{0.011}                                        & \second{\mstd{0.945}{0.010}}
                       & \mstd{0.916}{0.019}                                        & \mstd{0.910}{0.005}
                       & 0.843
                       & \mstd{0.738}{0.008}                                        & \mstd{0.801}{0.003}                                     & - \\
      SCALEX           & 2022                                                       & Deep Learning
                       & \mstd{0.673}{0.012}                                        & \mstd{0.924}{0.022}
                       & \mstd{0.942}{0.018}                                        & \mstd{0.928}{0.014}
                       & 0.867
                       & \second{\mstd{0.878}{0.002}}                               & \mstd{0.871}{0.009}                                     & - \\
      sysVI            & 2024                                                       & Deep Learning
                       & \best{\mstd{0.759}{0.023}}                                 & \mstd{0.911}{0.008}
                       & \mstd{0.931}{0.003}                                        & \mstd{0.925}{0.011}
                       & \second{0.882}
                       & \mstd{0.853}{0.005}                                        & \mstd{0.870}{0.003}                                     & - \\
      SCEMENT          & 2025                                                       & Traditional
                       & \mstd{0.658}{0.000}                                        & \mstd{0.883}{0.000}
                       & \mstd{0.840}{0.001}                                        & \mstd{0.909}{0.002}
                       & 0.823
                       & \best{\mstd{0.907}{0.005}}                                 & \mstd{0.856}{0.002}                                     & - \\
      scCobra          & 2025                                                       & Deep Learning
                       & \mstd{0.661}{0.001}                                        & \best{\mstd{0.953}{0.001}}
                       & \mstd{0.951}{0.001}                                        & \mstd{0.937}{0.001}
                       & 0.876
                       & \mstd{0.877}{0.006}                                        & \second{\mstd{0.876}{0.003}}                            & - \\
      \rowcolor{blue!3}
      \textbf{\method} & \textbf{Ours}                                              & Deep Learning
                       & \second{\mstd{0.756}{0.014}}                               & \mstd{0.890}{0.003}
                       & \best{\mstd{0.974}{0.001}}                                 & \best{\mstd{0.963}{0.002}}
                       & \best{0.896}
                       & \mstd{0.875}{0.002}                                        & \best{\mstd{0.887}{0.003}}                              & \textbf{+1.26\%} \\

      \addlinespace[0.4em]
      \rowcolor{gray!10}
      \multicolumn{11}{c}{\textbf{Immune (Human)}}                                                                                            \\
      scVI             & 2018                                                       & Deep Learning
                       & \mstd{0.536}{0.001}                                        & \mstd{0.823}{0.012}
                       & \mstd{0.799}{0.020}                                        & \mstd{0.809}{0.003}
                       & 0.742
                       & \mstd{0.921}{0.001}                                        & \mstd{0.814}{0.005}                                     & - \\
      Harmony          & 2019                                                       & Traditional
                       & \mstd{0.561}{0.000}                                        & \mstd{0.792}{0.000}
                       & \second{\mstd{0.814}{0.015}}                               & \best{\mstd{0.816}{0.003}}
                       & 0.746
                       & \mstd{0.920}{0.000}                                        & \second{\mstd{0.815}{0.003}}                            & - \\
      INSCT            & 2021                                                       & Deep Learning
                       & \mstd{0.561}{0.011}                                        & \mstd{0.886}{0.000}
                       & \mstd{0.581}{0.030}                                        & \mstd{0.737}{0.008}
                       & 0.691
                       & \mstd{0.813}{0.009}                                        & \mstd{0.740}{0.007}                                     & - \\
      SCALEX           & 2022                                                       & Deep Learning
                       & \second{\mstd{0.574}{0.002}}                               & \second{\mstd{0.915}{0.014}}
                       & \mstd{0.736}{0.025}                                        & \mstd{0.775}{0.011}
                       & 0.750
                       & \mstd{0.871}{0.001}                                        & \mstd{0.798}{0.006}                                     & - \\
      sysVI            & 2024                                                       & Deep Learning
                       & \best{\mstd{0.619}{0.008}}                                 & \mstd{0.855}{0.008}
                       & \mstd{0.779}{0.010}                                        & \mstd{0.745}{0.017}
                       & 0.750
                       & \mstd{0.899}{0.001}                                        & \mstd{0.810}{0.002}                                     & - \\
      SCEMENT          & 2025                                                       & Traditional
                       & \mstd{0.558}{0.000}                                        & \mstd{0.741}{0.000}
                       & \mstd{0.763}{0.012}                                        & \mstd{0.800}{0.006}
                       & 0.716
                       & \mstd{0.881}{0.000}                                        & \mstd{0.782}{0.003}                                     & - \\
      scCobra          & 2025                                                       & Deep Learning
                       & \mstd{0.563}{0.001}                                        & \best{\mstd{0.930}{0.003}}
                       & \mstd{0.730}{0.022}                                        & \mstd{0.779}{0.013}
                       & 0.751
                       & \mstd{0.878}{0.004}                                        & \mstd{0.801}{0.006}                                     & - \\
      \rowcolor{blue!3}
      \textbf{\method} & \textbf{Ours}                                              & Deep Learning
                       & \mstd{0.573}{0.003}                                        & \mstd{0.826}{0.006}
                       & \best{\mstd{0.816}{0.002}}                                 & \second{\mstd{0.810}{0.001}}
                       & \best{0.756}
                       & \second{\mstd{0.924}{0.002}}                               & \best{\mstd{0.823}{0.002}}                              & \textbf{+0.98\%} \\

      \bottomrule
    \end{tabular}%
  }
\end{table*}

\textbf{Dual-stream sparse diffusion 
(\Cref{fig:architecture}(B)).}
To capture the non-linear manifold structure of cells that is often missed by linear encoders, we employ a dual-stream architecture where each branch learns a structural representation via graph diffusion. Unlike generative diffusion, we use diffusion in the graph-signal-processing sense (feature propagation). For each stream \(*\in\{\VAR,\INV\}\), we learn a sparse adjacency matrix \(\mathbf{P}_*\) via differentiable query/key attention~\citep{vaswani2017attention} \(\mathbf{S}_*=\ReLU(\mathbf{Q}_*\mathbf{K}_*^\top/\tau)\) with Top-\(k\) sparsification.
Crucially, these are \emph{feature-level} (gene--gene) graphs of dimension \(G_*\times G_*\) (where each stream contains at most \(G_{\mathrm{sel}}=2{,}000\) genes), \emph{not} cell-level graphs.
This keeps the adjacency computation at \(\mathcal{O}(G_*^2)\)—independent of the number of cells \(N\) and orders of magnitude cheaper than building \(N\times N\) cell graphs.
Cell-level computation reduces to the sparse matrix product \(\mathbf{X}_*\,\mathbf{P}_*\), which scales \emph{linearly} in \(N\) (see Appendix~\ref{app:computational} for details).

\textbf{Stop-gradient caching for stability.}
Differentiating through iterative Top-\(k\) selection and multi-hop powers is unstable and computationally expensive. We therefore adopt a \emph{piecewise-constant topology} strategy: we periodically rebuild \(\mathbf{P}_*\) and cache its powers. During representation learning, we treat the graph structure as fixed (\(\sg(\mathbf{P}_*)\)), allowing gradients to update feature extractors without high-variance structural noise.

\textbf{Multi-scale extraction.}
We encode the gated variants \(\mathbf{X}_{\VAR}=\tilde{\mathbf{X}}_{\VAR}\) and raw anchors \(\mathbf{X}_{\INV}\) separately. For scale \(k\in\mathcal{K}\), we compute low-pass features (smoothing global structure) and high-pass features (sharpening local details):
\begin{equation}
    \label{eq:diffusion_terms}
    \begin{aligned}
        \mathbf{Z}_*^{(k),\mathrm{low}}
         & \coloneqq
        \mathbf{X}_*\,\sg(\mathbf{P}_*)^{k},
        \\
        \mathbf{Z}_*^{(k),\mathrm{high}}
         & \coloneqq
        \xi_1\bigl(\mathbf{X}_*-\mathbf{Z}_*^{(k),\mathrm{low}}\bigr)
        +
        \xi_2\,\mathbf{X}_*\bigl(\mathbf{I}-\sg(\mathbf{P}_*)^{m}\bigr).
    \end{aligned}
\end{equation}
Scales are aggregated via learnable weights \(\omega=\softmax(\theta)\) to produce stream embeddings \(\mathbf{H}_{\VAR}\) and \(\mathbf{H}_{\INV}\). To prevent collapse before interaction, each stream is regularized by a decoder reconstructing its own input. We formally prove that this multi-scale diffusion acts as a complementary spectral filter (preserving global structure while isolating local details) in Appendix~\ref{app:theory_spectral}.

\begin{figure*}[t]
    \centering
    \includegraphics[width=0.99\linewidth]{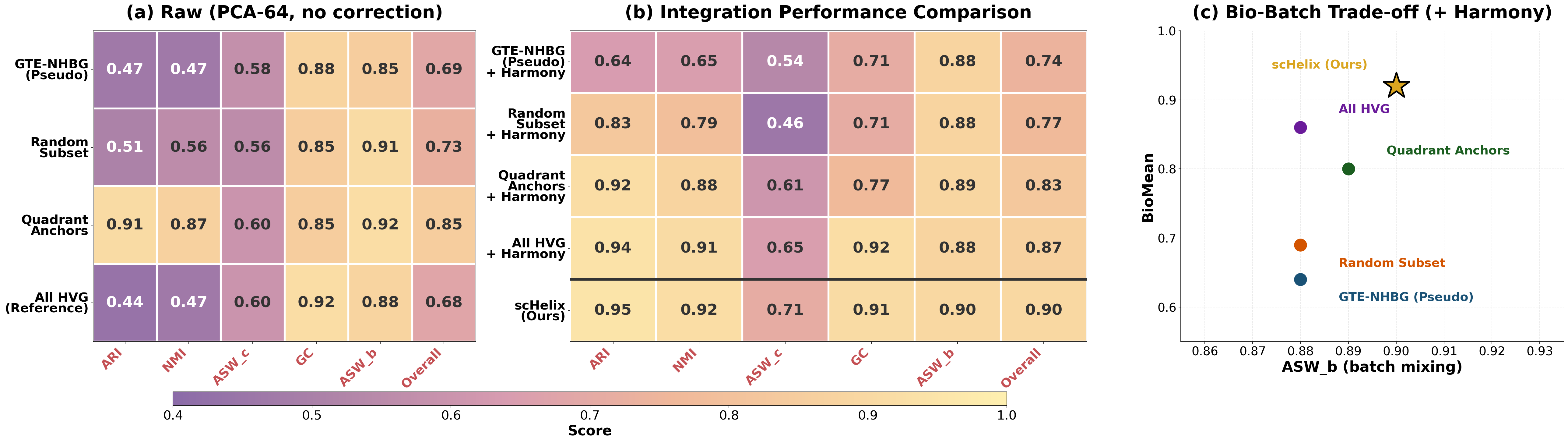}
    \caption{\textbf{Anchor-set diagnostics on Human Pancreas (complexBatch).}
        (a) Metric heatmaps on raw PCA-64 embeddings show Quadrant Anchors provide the best initial biological separation.
        (b) Performance after Harmony integration.
        (c) The bio-batch trade-off demonstrates that while Anchors are stable, fusing all genes (Ours/scHelix) achieves the optimal balance.}
    \label{fig:raw_vs_harmony_heatmap_tradeoff}
    \Description{Three-panel figure showing anchor-set diagnostics: panel a displays metric heatmaps on raw PCA-64 embeddings for different gene subsets; panel b shows the same metrics after Harmony correction; panel c plots the bio-batch trade-off curve comparing different gene subsets with the full scHelix method achieving the best balance.}
    \vspace{-1em}
\end{figure*}

\subsection{Align-Refine Interaction \& HyperFusion}
\label{sec:serial_fusion}

\textbf{Design principle 
(\Cref{fig:architecture}(C)).}
Naive fusion can reintroduce domain artifacts into the clean Anchor scaffold. To prevent this, scHelix enforces a strict \emph{Align-Refine-Fuse} protocol, designed to \emph{prevent noise leakage} from the unstable Variant stream back to the stable Anchor topology. The process proceeds as follows: (i) align the Variant stream to the Anchor stream, (ii) conservatively refine the Anchor stream using denoised Variant details, and (iii) fuse via a gated mechanism.

\textbf{Asymmetric Variant--Anchor Alignment.}
We align the unstable Variant stream to the robust Anchor topology using a BYOL/SimSiam-style objective \citep{grill2020byol, chen2021simsiam}. We minimize the negative cosine similarity between the Variant stream's prediction \(\mathbf{p}_{\mathrm{S}}\) and the Anchor stream's target \(\mathbf{z}_{\mathrm{T}}\):
\begin{equation}
    \label{eq:align_loss}
    \mathcal{L}_{\mathrm{align}}
    \coloneqq
    2-2\,\E\!\left[\left\langle\mathbf{p}_{\mathrm{S}},\mathbf{z}_{\mathrm{T}}\right\rangle\right].
\end{equation}
Crucially, the stop-gradient on \(\mathbf{z}_{\mathrm{T}}\) ensures the Anchor manifold acts as a fixed topological guide, preventing it from being pulled toward the noisy Student distribution (i.e., preventing model collapse).

\textbf{Conservative Reverse Refinement.}
Once aligned, the Student stream contains recovered biological signals. We inject these back into the Teacher stream via a \emph{bounded residual update}:
\begin{equation}
    \label{eq:refine_update}
    \tilde{\mathbf{H}}_{i}^{\INV} \coloneqq \mathbf{H}_{i}^{\INV} + \boldsymbol{\alpha}_i \odot \Delta\mathbf{h}_i, \qquad \mathbf{0} \le \boldsymbol{\alpha}_i \le \alpha_{\max}.
\end{equation}
where \(\Delta\mathbf{h}_i\) is derived from cross-attention (Anchor queries Variant), and \(\boldsymbol{\alpha}_i\) is an element-wise gating vector. The bound \(\alpha_{\max}\) acts as a safety valve (theoretically guaranteed by Lipschitz continuity, see Appendix~\ref{app:theory_boundedness}), ensuring that while the Anchor stream gains detail, its global structure cannot be destabilized by residual batch noise.

\textbf{HyperFusion.}
Finally, we synthesize the refined streams using \emph{HyperFusion}.
A HyperNetwork~\citep{ha2016hypernetworks} generates instance-conditioned low-rank parameters \(\boldsymbol{\theta}_i\) from the refined Anchor embedding \(\tilde{\mathbf{H}}_i^{\INV}\); these parameters define a two-layer MLP that transforms the raw Variant embedding \(\mathbf{h}_i^{\VAR}\) into a residual update.
An adaptive gate \(\mathbf{g}_i=\sigma(\boldsymbol{\theta}_i^{\mathrm{gate}})\) controls the injection magnitude:
\begin{equation}
    \begin{aligned}
        \label{eq:hyperfuse}
        \boldsymbol{\theta}_i &\coloneqq \mathrm{HyperNet}\bigl(\tilde{\mathbf{H}}_i^{\INV}\bigr),\\
        \mathbf{H}_{i,\mathrm{fused}}
        &\coloneqq
        \LN\bigl(
        \tilde{\mathbf{H}}_{i}^{\INV}
        +\mathbf{g}_i \odot \lambda_\Delta\,
        \mathrm{MLP}(\mathbf{H}_i^{\VAR};\,\boldsymbol{\theta}_i)
        \bigr).
    \end{aligned}
\end{equation}
This adaptive gating allows the model to selectively incorporate Variant information—conditioned on the local Anchor context—only where it is biologically congruent, preventing residual batch noise from contaminating the fused representation.

\subsection{Optimization \& Teacher-Guided Schedule}
\label{sec:objective}

We employ a two-phase curriculum to ensure stability.

\textbf{Phase 1: Warm-up (Decoupled).}
Initially, we train the streams with a focus on reconstruction and asymmetric alignment. The objective includes per-stream reconstruction (\(\mathcal{L}_{\mathrm{rec}}\)), alignment (\(\mathcal{L}_{\mathrm{align}}\)), and topology preservation (\(\mathcal{L}_{\mathrm{conn}}\)), $\mathcal{L}_{\mathrm{warm}} \coloneqq $
\begin{equation}
    \label{eq:warm_obj}
    \sum_{*\in\{\VAR,\INV\}} \left( \lambda^{*}_{\mathrm{rec}}\|\hat{\mathbf{X}}_{*}-\mathbf{X}_{*}\|_F^2 + \lambda^{*}_{\mathrm{conn}}\mathcal{L}_{\mathrm{conn}}(\mathbf{H}_{*}) \right)
    +\lambda_{\mathrm{align}}\mathcal{L}_{\mathrm{align}}.
\end{equation}
During this phase, we also maintain Teacher prototypes \(\{\mathbf{c}_k\}\) via EMA~\citep{tarvainen2017mean} to generate stable pseudo-labels \(y_i\) for the connectivity loss.

\textbf{Phase 2: Fusion (Coupled).}
After \(T_{\mathrm{warm}}\) steps, we activate the fusion module. The model minimizes a joint objective that includes fused reconstruction and \emph{Confidence-Weighted Distillation}~\citep{hinton2015distilling} (\(\mathcal{L}_{\mathrm{KD}}\)), which encourages the fused embedding to respect the high-confidence cluster structure found by the refined Teacher:
\begin{equation}
    \label{eq:fuse_obj}
    \mathcal{L}_{\mathrm{fuse}} \coloneqq
    \lambda^{\mathrm{fused}}_{\mathrm{rec}}\|\hat{\mathbf{X}}-\mathbf{X}\|_F^{2}
    +\lambda_{\mathrm{KD}}\mathcal{L}_{\mathrm{KD}}
    +\lambda^{\mathrm{fused}}_{\mathrm{conn}}\mathcal{L}_{\mathrm{conn}}(\mathbf{H}_{\mathrm{fused}}).
\end{equation}
The total objective is \(\mathcal{L} \coloneqq \mathcal{L}_{\mathrm{warm}} + \mathbb{I}_{\mathrm{fuse}}(t)\,\mathcal{L}_{\mathrm{fuse}}\), ensuring that fusion only occurs after the Anchor scaffold and Variant alignment are sufficiently mature.

\section{Experiments}
\label{sec:experiments}
\subsection{Experimental Setup}
\label{sec:exp_setup}

\paragraph{\textbf{Task, datasets, baselines, and metrics.}}
We study multi-domain scRNA-seq integration as representation learning under domain shift~\citep{gong2016domain}.
We benchmark five public datasets; the main text reports three representative ones
(Table~\ref{tab:main_benchmark_compact_yearsorted}), with the remaining two deferred to the Appendix
(Table~\ref{tab:deep_3seed_comparison}).
Baselines include Harmony, scVI, INSCT, SCALEX, sysVI, SCEMENT, and scCobra
(\citealp{korsunsky2019harmony,lopez2018deep,simon2021insct,xiong2022scalex,sysvi2024,chockalingam2025scement,sccobra2025}).
All methods share the same QC/HVG pool and a unified training/evaluation pipeline (Appendix~\ref{app:preproc}--\ref{app:train_eval}).
Following scIB conventions~\citep{luecken2022benchmarking}, all metrics are scaled/flipped to $[0,1]$ (higher is better).
We report the standard scIB Overall score (40/60 batch/bio weighting); detailed metric definitions are provided in \Cref{tab:main_benchmark_compact_yearsorted}.

\subsection{Main Integration Performance}
\label{sec:exp_main_results}

\begin{figure*}[t]
  \centering
  \includegraphics[width=0.98\textwidth]{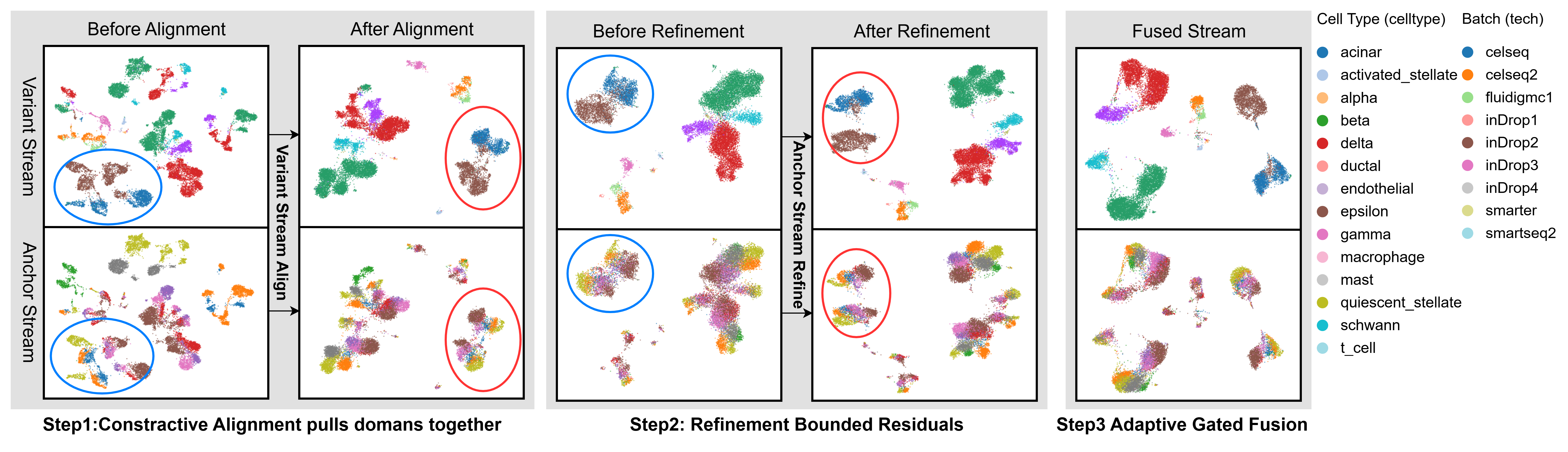}
  \caption{\textbf{Align-Refine-Fuse trajectory} on Human Pancreas (complexBatch).
    \textbf{(A)} Alignment pulls the Variant stream toward the Anchor manifold, mixing batches.
    \textbf{(B)} Refinement injects denoised details into the Anchor, sharpening cluster boundaries.
    \textbf{(C)} Fusion yields the final embedding with compact biological clusters and minimal batch stratification.
    \textcolor{red}{Red} circles highlight improved local compactness by \textsc{scHelix}.}
  \label{fig:interaction_umap}
  \Description{Three-panel UMAP visualization showing the Align-Refine-Fuse trajectory on Human Pancreas data: panel A shows alignment mixing batches, panel B shows refinement sharpening cluster boundaries, and panel C shows the final fused embedding with compact clusters and minimal batch stratification.}
  \vspace{-0.3cm}
\end{figure*}

\begin{figure*}[t]
  \centering
  \includegraphics[width=1.02\textwidth]{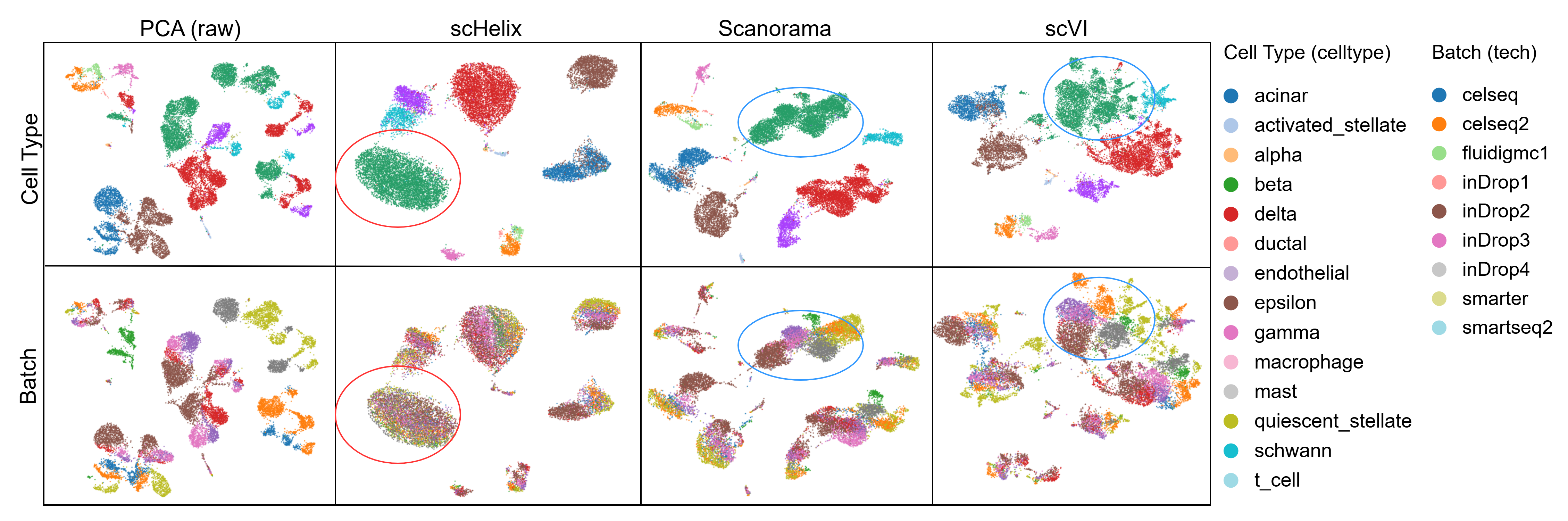}
  \vspace{-2.4em}
  \caption{\textbf{Comparative UMAP plots of four models on Human Pancreas (complexBatch).}
    Red circles highlight improved local compactness by \method; blue circles highlight baseline defects,
    including diffuse neighborhoods and reduced separation between ductal and beta populations due to over-correction.}
  \label{fig:umap_method_comparison_pears}
  \Description{Comparative UMAP embeddings of four integration models on Human Pancreas, with red circles marking improved local compactness by scHelix and blue circles marking baseline defects such as diffuse neighborhoods and over-merged ductal and beta populations.}
  \vspace{-0.4em}
\end{figure*}

\paragraph{\textbf{Benchmark results.}}
Table~\ref{tab:main_benchmark_compact_yearsorted} summarizes three representative datasets
(full five-dataset results with mean$\pm$std over three seeds: Appendix Table~\ref{tab:deep_3seed_comparison}).
Across all three, \method achieves the best Overall score, surpassing the second-best baseline by an average of 1.05\%.
Gains are most pronounced in distinguishing subtle cell states.
On the Pancreas dataset, \textbf{Harmony and scVI tend to over-merge Ductal and Acinar populations}
(\Cref{fig:umap_method_comparison_pears}, red circles) in pursuit of batch removal,
whereas \method preserves these local neighborhoods while maintaining comparable batch integration scores.
\textit{This structural preservation is underpinned by molecular fidelity:}
\method maintains sharp expression boundaries of lineage-specific markers
(e.g., \textit{KRT19} for Ductal vs.\ \textit{PRSS1} for Acinar~\citep{muraro2016single}),
avoiding the ``expression smoothing'' common in VAE-based baselines where marker specificity is diluted by the Gaussian latent prior.
This confirms that our asymmetric design effectively prevents over-correction, ensuring fine-grained biological identities are not sacrificed for batch removal.

\subsection{Mechanism Analysis}
\label{sec:exp_mechanism}

\paragraph{\textbf{Ablation study.}}
\label{sec:ablation}
We validate each component by systematically removing it (\Cref{tab:ablation_main});
the most diagnostic dataset (Human Pancreas, complexBatch)~\cite{luecken2022benchmarking} is reported here,
with Lung results in Appendix~\ref{app:supplementary_results}.
Three key findings emerge:
(i) \textbf{Anchor choice is most critical:} Replacing Quadrant selection with random anchors (S1) causes the sharpest drop (Overall \(0.891 \rightarrow 0.763\)), confirming that biologically discriminative anchors are essential.
(ii) \textbf{Interaction is necessary:} Disabling the Align-Refine protocol (S3) degrades the bio-batch trade-off, with specific drops in batch mixing when Alignment is removed.
(iii) \textbf{Teacher guidance stabilizes fusion:} Removing self-distillation (S5) yields a significant drop (Overall \(0.891 \rightarrow 0.788\)), underscoring the importance of a stable teacher topology.

\begin{table}[t]
  \centering
  \caption{\textbf{Ablation study on Human Pancreas.} We compare \textbf{S0} (Full) with variants: \textbf{S1} (random anchors), \textbf{S2} (no gating), \textbf{S3} (no interaction; \textbf{S3.1} no align, \textbf{S3.2} no refine), \textbf{S4} (simple fusion), \textbf{S5} (no self-training; \textbf{S5.1} no KD, \textbf{S5.2} no connectivity), and \textbf{S6} (linear encoder).}
  \label{tab:ablation_main}

  \begingroup
  \small
  \sisetup{round-precision=1}
  \renewcommand{\arraystretch}{1.1}
  \setlength{\tabcolsep}{6pt}

  \resizebox{\linewidth}{!}{
  \begin{tabular}{l
      S[table-format=2.1]
      S[table-format=2.1]
      S[table-format=2.1]
      S[table-format=2.1]
      S[table-format=2.1]
      S[table-format=2.1]}
    \toprule
    \textbf{Setting} &
    \multicolumn{4}{c}{\textbf{Bio metrics} (\(\uparrow\))} &
    \multicolumn{1}{c}{\textbf{Batch}} &
    \multicolumn{1}{c}{\textbf{Total}} \\
    \cmidrule(lr){2-5}\cmidrule(lr){6-6}\cmidrule(lr){7-7}
    & \multicolumn{2}{c}{\textbf{Clustering}} & %
    \multicolumn{2}{c}{\textbf{Structure}} & %
    \multicolumn{1}{c}{(\(\uparrow\))} &
    \multicolumn{1}{c}{(\(\uparrow\))} \\
    \cmidrule(lr){2-3}\cmidrule(lr){4-5}
    & \multicolumn{1}{c}{\textbf{ARI}} &
    \multicolumn{1}{c}{\textbf{NMI}} &
    \multicolumn{1}{c}{\textbf{ASW\(_{\mathrm{ct}}\)}} &
    \multicolumn{1}{c}{\textbf{GC}} &
    \multicolumn{1}{c}{\textbf{ASW\(_{\mathrm{b}}\)}} & %
    \multicolumn{1}{c}{\textbf{Overall}} \\
    \midrule

    \rowcolor{gray!10}
    \textbf{S0 Full}             & \bfseries 93.1 & \bfseries 89.1 & \bfseries 74.4 & 92.3 & 92.0 & \bfseries 89.1 \\
    S1 RandSplit               & 53.3 & 74.8 & 60.7 & 82.0 & 89.1 & 76.3 \\
    S2 NoGate                    & 93.0 & 88.9 & 71.2 & 90.4 & 90.0 & 87.9 \\
    S3 NoInteract             & 74.8 & 83.6 & 68.0 & 90.7 & 91.0 & 84.0 \\
    \hspace{1em}S3.1 NoAlign     & 90.2 & 87.0 & 56.1 & 92.8 & 91.4 & 85.4 \\
    \hspace{1em}S3.2 NoRefine    & 92.5 & 88.5 & 71.8 & 82.3 & 91.2 & 86.8 \\
    S4 SimpFusion              & 91.6 & 86.8 & 54.9 & 92.0 & \bfseries 93.6 & 86.2 \\
    S5 NoSelfTrain            & 52.6 & 75.6 & 56.4 & 93.0 & 93.0 & 78.8 \\
    \hspace{1em}S5.1 NoKD        & 91.8 & 86.7 & 57.6 & 85.5 & 91.9 & 85.0 \\
    \hspace{1em}S5.2 NoConn      & 56.0 & 77.8 & 58.2 & \bfseries 93.2 & 92.2 & 79.7 \\
    S6 LinEncoder             & 91.3 & 87.3 & 69.7 & 89.5 & 91.4 & 87.2 \\
    \bottomrule
  \end{tabular}
  }
  \endgroup
  \vspace{-1em}
\end{table}

\paragraph{\textbf{Diagnostic I: Anchor Set Quality.}}
\label{sec:anchor_set_diagnostics}
We benchmark the intrinsic quality of candidate gene subsets prior to deep integration.
Four subsets drawn from the same HVG pool are compared:
(i) \textbf{GTE-NHBG} (GTE \(\ge 0.95\)~\citep{zhou2025gte});
(ii) \textbf{Quadrant Anchors} (our selection: \(B_z(g)\le 0,\, C_z(g)\ge 0\));
(iii) \textbf{All HVGs}; and
(iv) \textbf{Random subsets} (size-matched, averaged over \(R\) repeats).
For each set, we derive PCA-64 embeddings from scaled expression and evaluate clustering
(\(\mathrm{ARI}_{\mathrm{best}}\), \(\mathrm{NMI}_{\mathrm{best}}\)) and conservation metrics
(\(\mathrm{ASW}_{\mathrm{ct}}\), GC, \(\mathrm{ASW}_{\mathrm{batch}}\)) before and after Harmony~\citep{korsunsky2019harmony}.
As shown in \Cref{fig:raw_vs_harmony_heatmap_tradeoff}, \textbf{Quadrant Anchors} form the most robust biological scaffold in raw space (Panel a),
yet post-correction results (Panels b, c) show that incorporating Variants further enhances resolution.
This empirically motivates our asymmetric design: Anchors provide topological stability,
while controlled fusion recovers the fine-grained detail needed for high-resolution clustering.

\paragraph{\textbf{Diagnostic II: Interaction Mechanism.}}
\label{sec:interaction_diagnosis}
We quantify biological conservation (BioMean) on intermediate embeddings to dissect the asymmetric protocol.
The net gain (\(\Delta\)BioMean) is measured for:
(i) the Variant stream via \textbf{Align} (S3.1 vs.\ S0), and
(ii) the Anchor stream via \textbf{Refine} (S3.2 vs.\ S0).
\Cref{tab:interaction_effect_compact} confirms that \textbf{Align} substantially boosts the noisy Variant stream by anchoring it to the stable Teacher topology,
while \textbf{Refine} yields large gains on the Anchor branch by injecting bounded residuals without compromising structural integrity.
This synergy is visually corroborated in \Cref{fig:interaction_umap}: Alignment mitigates batch offsets, Refinement sharpens local structures, and the fused embedding balances rigorous batch removal with high-fidelity biological conservation.

\begin{table}[t]
\centering
\caption{Branch-level diagnostics for Align and Refine. We report BioMean on branch embeddings and the net improvement (\(\Delta\)) provided by each module. Bold indicates the best performance per setting.}
\label{tab:interaction_effect_compact}

\begingroup
\small
\sisetup{round-precision=1}
\renewcommand{\arraystretch}{1.1}
\setlength{\tabcolsep}{0pt}

\begin{tabular*}{\linewidth}{@{\extracolsep{\fill}} l l 
  S[table-format=2.1] 
  S[table-format=2.1] 
  S[table-format=2.1] 
  S[table-format=2.1] 
  S[table-format=2.1] 
  c @{}}
\toprule
\multirow{3}{*}{\textbf{Setting}} & \multirow{3}{*}{\textbf{Stream}} & \multicolumn{4}{c}{\textbf{Bio metrics ($\uparrow$)}} & \multicolumn{2}{c}{\textbf{Overall ($\uparrow$)}} \\
\cmidrule(lr){3-6} \cmidrule(lr){7-8}
 & & \multicolumn{2}{c}{\textbf{Cluster}} & \multicolumn{2}{c}{\textbf{Structure}} & \multicolumn{1}{c}{\multirow{2}{*}{\textbf{BioMean}}} & \multirow{2}{*}{\textbf{$\Delta$}} \\
\cmidrule(lr){3-4} \cmidrule(lr){5-6}
 & & \multicolumn{1}{c}{\textbf{ARI}} & \multicolumn{1}{c}{\textbf{NMI}} & \multicolumn{1}{c}{\textbf{ASW$_{ct}$}} & \multicolumn{1}{c}{\textbf{GC}} & & \\
\midrule

\multicolumn{8}{l}{\cellcolor{gray!10}\textbf{Dataset: Lung\_atlas\_public}} \\
\multirow{2}{*}{S3.1 (Contrast)} & HBG (w/o Align) & 56.2 & 73.2 & 55.4 & 83.9 & 67.2 & -- \\
                                 & HBG (w/ Align)  & \bfseries 58.9 & \bfseries 75.0 & \bfseries 57.1 & \bfseries 84.9 & \bfseries 69.0 & \bfseries +1.8 \\
\addlinespace[0.3em]
\multirow{2}{*}{S3.2 (Refine)}   & NHBG (w/o Refine) & 47.2 & 70.9 & \bfseries 57.1 & 82.7 & 64.5 & -- \\
                                 & NHBG (w/ Refine)  & \bfseries 58.9 & \bfseries 75.0 & \bfseries 57.1 & \bfseries 84.9 & \bfseries 69.0 & \bfseries +4.5 \\

\midrule

\multicolumn{8}{l}{\cellcolor{gray!10}\textbf{Dataset: Human Pancreas (complexBatch)}} \\
\multirow{2}{*}{S3.1 (Contrast)} & HBG (w/o Align) & 90.2 & 87.0 & 56.1 & 92.8 & 81.5 & -- \\
                                 & HBG (w/ Align)  & \bfseries 93.1 & \bfseries 89.1 & \bfseries 74.4 & 92.3 & \bfseries 87.2 & \bfseries +5.7 \\
\addlinespace[0.3em]
\multirow{2}{*}{S3.2 (Refine)}   & NHBG (w/o Refine) & 92.5 & 88.5 & 71.8 & 82.3 & 83.8 & -- \\
                                 & NHBG (w/ Refine)  & \bfseries 93.1 & \bfseries 89.1 & \bfseries 74.4 & \bfseries 92.3 & \bfseries 87.2 & \bfseries +3.5 \\

\bottomrule
\end{tabular*}
\endgroup
\end{table}

\subsection{Extended Capabilities \& Robustness}
\label{sec:exp_extended}

\begin{table}[t]
    \centering
    \caption{\textbf{Semi-supervised integration on Human Pancreas.} Comparison against scANVI across varying label ratios (10\%--100\%). \(^{\dagger}\) semi-supervised; \(^{\ddagger}\) fully supervised. Bold indicates best performance.}
    \label{tab:semi_supervised}
    \begingroup
    \small
    \sisetup{round-precision=1}
    \renewcommand{\arraystretch}{1.1}
    \setlength{\tabcolsep}{0pt}
    \begin{tabular*}{\linewidth}{@{\extracolsep{\fill}} l l
            S[table-format=2.1]
            S[table-format=2.1]
            S[table-format=2.1]
            S[table-format=2.1]
            S[table-format=2.1]
            c @{}}
        \toprule
        & & \multicolumn{4}{c}{\textbf{Bio metrics} ($\uparrow$)} & \multicolumn{2}{c}{\textbf{Overall} ($\uparrow$)} \\
        \cmidrule(lr){3-6} \cmidrule(lr){7-8}
        & & \multicolumn{2}{c}{\textbf{Cluster}} & \multicolumn{2}{c}{\textbf{Structure}} & \multicolumn{1}{c}{\multirow{2}{*}{\textbf{Batch}}} & \multirow{2}{*}{\textbf{Score}} \\
        \cmidrule(lr){3-4} \cmidrule(lr){5-6}
        \textbf{Ratio} & \textbf{Method} & \multicolumn{1}{c}{\textbf{ARI}} & \multicolumn{1}{c}{\textbf{NMI}} & \multicolumn{1}{c}{\textbf{ASW}$_{ct}$} & \multicolumn{1}{c}{\textbf{GC}} & & \\
        \midrule
        \multirow{2}{*}{\textbf{10\%}}
        & scANVI$^{\dagger}$ & 95.0 & 91.2 & 65.3 & 87.6 & 88.1 & 86.1 \\
        & \method$^{\dagger}$ & \bfseries 96.8 & \bfseries 93.9 & \bfseries 78.8 & \bfseries 92.2 & \bfseries 91.7 & \bfseries 90.9 \\
        \addlinespace[0.3em]
        \multirow{2}{*}{\textbf{20\%}}
        & scANVI$^{\dagger}$ & 95.0 & 91.7 & 65.3 & 87.8 & 88.2 & 86.2 \\
        & \method$^{\dagger}$ & \bfseries 97.1 & \bfseries 94.5 & \bfseries 82.8 & \bfseries 94.6 & \bfseries 91.6 & \bfseries 92.0 \\
        \addlinespace[0.3em]
        \multirow{2}{*}{\textbf{50\%}}
        & scANVI$^{\dagger}$ & 95.2 & 92.0 & 65.4 & 89.0 & 88.5 & 86.7 \\
        & \method$^{\dagger}$ & \bfseries 98.9 & \bfseries 96.9 & \bfseries 84.9 & \bfseries 94.7 & \bfseries 90.8 & \bfseries 92.6 \\
        \addlinespace[0.3em]
        \multirow{2}{*}{\textbf{100\%}}
        & scANVI$^{\ddagger}$ & 96.1 & 93.4 & 65.4 & 90.1 & \bfseries 91.1 & 89.2 \\
        & \method$^{\ddagger}$ & \bfseries 99.0 & \bfseries 97.3 & \bfseries 86.9 & \bfseries 94.7 & 90.9 & \bfseries 93.1 \\
        \bottomrule
    \end{tabular*}
    \endgroup
\end{table}

\paragraph{\textbf{Semi-supervised setting and zero-shot foundation models.}}
\label{sec:exp_fm_and_semisup}
We extend evaluation to label-scarce scenarios (10--50\% masking).
Comparison with scANVI~\citep{xu2021probabilistic} (\Cref{tab:semi_supervised}) shows that \method consistently yields superior integration.
Even with just 10\% labels, \method improves Overall by \(+0.048\) (0.861\,\(\rightarrow\)\,0.909), driven by substantial \(\mathrm{ASW}_{\mathrm{ct}}\) gains (\(+0.13\)),
confirming that Teacher-guided topology effectively propagates sparse label information.

We further investigate whether foundation models (FMs)~\cite{scfoundation_survey2024} can replace dataset-adaptive integration,
benchmarking \textbf{Geneformer}~\citep{theodoris2023geneformer} and \textbf{scGPT-human}~\citep{scgpt2024} on \texttt{Human Pancreas}
with both raw and Harmony-integrated embeddings.

\begin{figure*}[t]
  \centering
  \includegraphics[width=\linewidth]{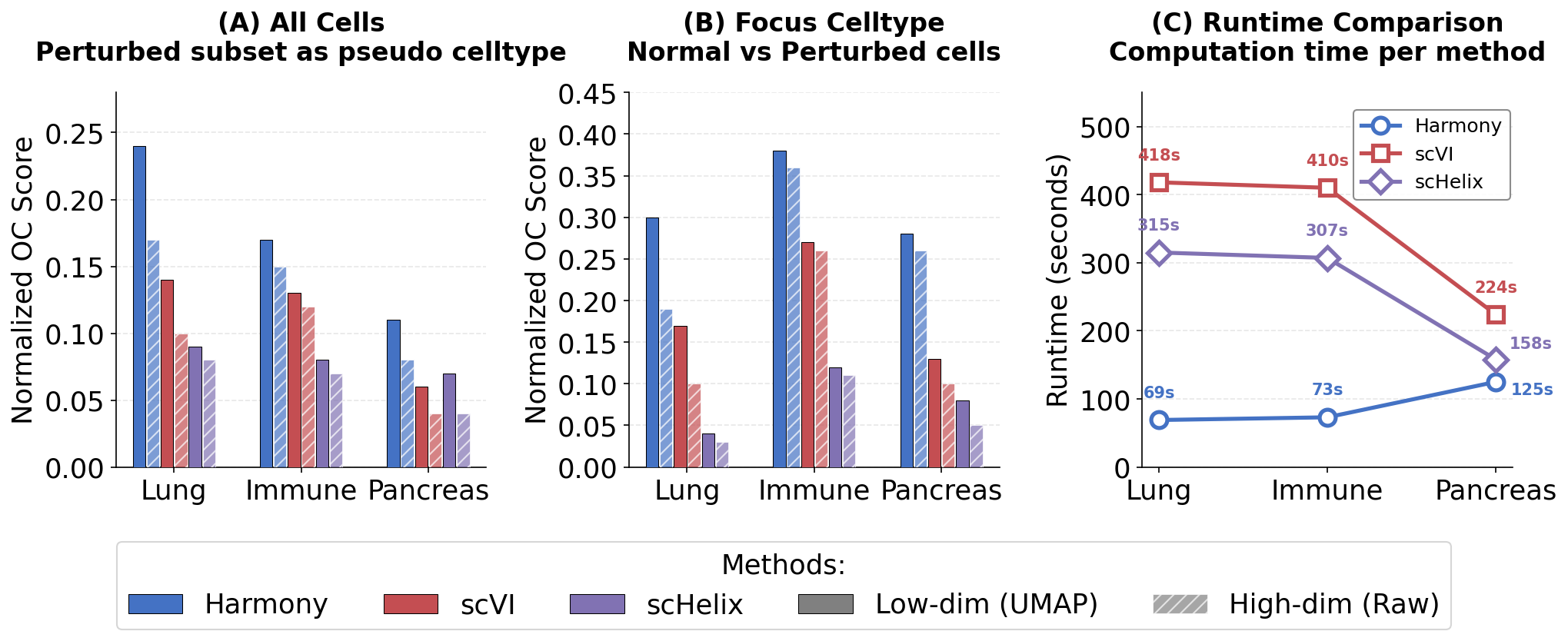}
  \vspace{-2.6em}
  \caption{\textbf{Over-correction and runtime.}
    Bars: Over-correction (lower is better) in high-dim vs.\ 2D space. Lines: Runtime.
    \method consistently achieves the lowest global (\textbf{A}, \(\text{OC}_{all}\)) and local (\textbf{B}, \(\text{OC}_{focus}\)) over-correction scores while maintaining computational efficiency.}
  \label{fig:overcorr_runtime}
  \Description{Bar-and-line chart showing over-correction scores and runtime for multiple integration methods across datasets: scHelix consistently achieves the lowest global and local over-correction scores while maintaining runtime comparable to lightweight methods.}
\end{figure*}

\begin{figure*}[t]
  \centering
  \includegraphics[width=\linewidth]{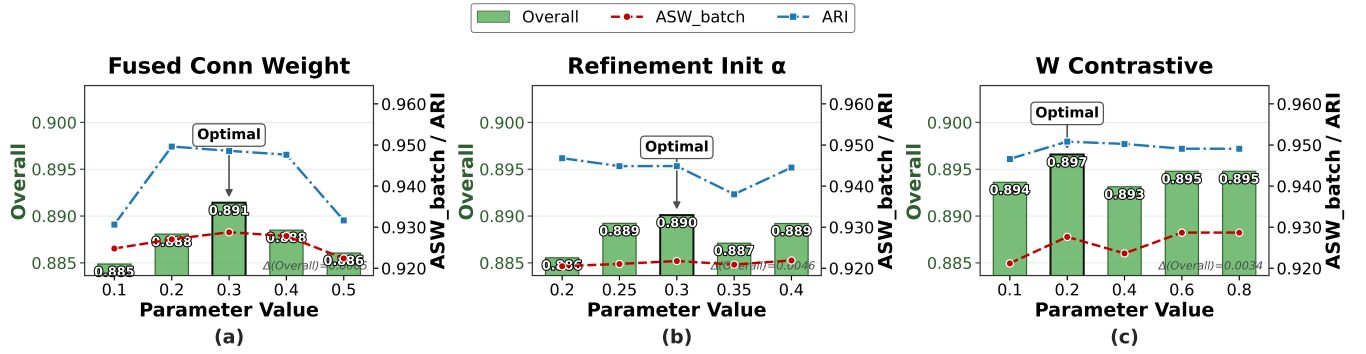}
  \vspace{-2.6em}
  \caption{\textbf{Hyperparameter sensitivity on Pancreas (complexBatch).}
    One knob is varied at a time with all others fixed.
    Bars: Overall; dashed curves: \(\mathrm{ASW}_{\mathrm{batch}}\) and \(\mathrm{ARI}_{\mathrm{best}}\).
    Results shown for (a) \(\lambda_{\mathrm{fused\_conn}}\),
    (b) \(\alpha_{\mathrm{init}}\),
    and (c) \(\lambda_{\mathrm{align}}\) (denoted \(w_{\mathrm{contrastive}}\) in plots).}
  \label{fig:sensitivity_analysis}
  \Description{Three-panel bar-and-line chart showing hyperparameter sensitivity: each panel varies one hyperparameter while displaying Overall score as bars and ASW-batch and ARI-best as dashed curves, demonstrating stable performance across a broad range of settings.}
  \vspace{-0.6em}
\end{figure*}

\begin{figure}[t]
  \centering
  \includegraphics[width=\linewidth]{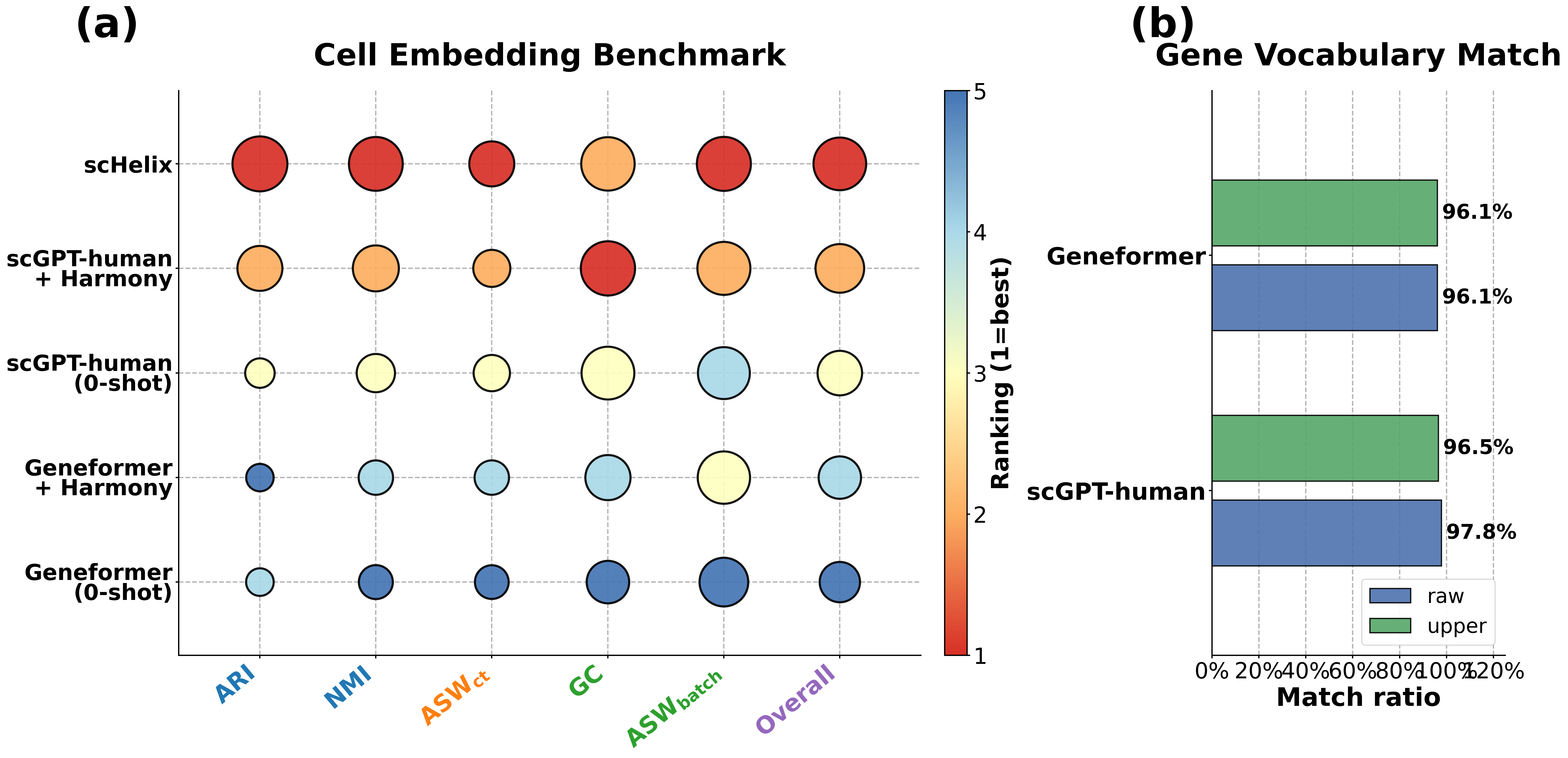}
  \vspace{-0.6cm}
  \caption{\textbf{Zero-shot FM comparison on Human Pancreas.} (\textbf{a}) scIB metrics comparing scHelix against Geneformer and scGPT (with/without Harmony). (\textbf{b}) Vocabulary match ratios indicate high feature coverage.}
  \label{fig:fm_zeroshot_comparison}
  \Description{Bar charts comparing zero-shot foundation model performance: panel a shows scIB metric scores for scHelix, Geneformer, and scGPT with and without Harmony integration; panel b shows vocabulary match ratios exceeding 96 percent for both foundation models.}
  \vspace{-0.3cm}
\end{figure}

\Cref{fig:fm_zeroshot_comparison} reveals that \method achieves a significantly better bio-batch trade-off (Overall: 0.895) than the best FM (scGPT+Harmony: 0.796).
Gene vocabulary coverage exceeds 96\% for both FMs, so the gap stems not from missing features but from the fact that FMs are pre-trained for reconstruction rather than batch invariance, preserving technical artifacts without explicit adaptation.

\paragraph{\textbf{Over-correction \& Runtime.}}
\label{sec:exp_visualization_overcorr}
We employ a rigorous over-correction stress test following scCobra~\citep{sccobra2025}.
As shown in \Cref{fig:overcorr_runtime}, \method achieves the lowest OC scores among deep baselines, indicating superior local-neighborhood preservation.
This performance does not sacrifice efficiency:
\method scales linearly with cell number, processing the 30k-cell Pancreas dataset in \textbf{\(\sim\)150\,s} with peak memory under 4\,GB on a single A100 GPU—comparable to lightweight Harmony (\textbf{\(\sim\)120\,s}) but significantly faster than scVI (\textbf{\(\sim\)600\,s}) and scGPT (\textbf{\(>\)2000\,s}).

\paragraph{\textbf{Sensitivity analysis.}}
We vary key hyperparameters on Human Pancreas (\texttt{complexBatch}).
\method exhibits high stability (\Cref{fig:sensitivity_analysis}), with the Overall score fluctuating negligibly (\(\Delta \leq 0.0065\)) across a broad range of settings, indicating that the default configuration is near-optimal and no precise tuning is required.

\section{Conclusion}
We presented \method, a framework that resolves the integration trade-off by explicitly disentangling input genes into stable Anchors and batch-sensitive Variants. Through a conservative Align-Refine-Fuse protocol, \method achieves precise batch correction while preserving subtle biological signals. Extensive benchmarks confirm its superiority over state-of-the-art methods and foundation models, establishing it as a robust tool for high-fidelity cell atlas construction.

\section{Limitations and Ethical Considerations}
\paragraph{\textbf{Limitations.}}
First, the graph-based architecture incurs higher costs than linear methods, potentially requiring optimization for multi-million cell datasets.
Second, our strategy relies on the existence of shared biological signals; extreme non-overlapping batch effects may hinder effective anchor discovery.

\paragraph{\textbf{Ethical Considerations.}}
As an AI tool for science, scHelix warrants careful use: the ``refined'' expression values are computational imputations and must not substitute for experimentally validated clinical diagnoses.
Users must also adhere to data privacy regulations when processing patient-derived transcriptomics to prevent re-identification risks.

\IfFileExists{sections/discussion.tex}{\input{sections/discussion}}{}
\IfFileExists{sections/conclusion.tex}{\input{sections/conclusion}}{}

\FloatBarrier

\IfFileExists{main_scHelix_arxiv.bbl}{%

%
}{%
  \PackageError{scHelix}{Missing main_scHelix_arxiv.bbl}{Upload main_scHelix_arxiv.bbl together with main_scHelix_arxiv.tex.}%
}

\newpage
\appendix

\ifdefined\SCHELIX_APPENDIX_LOADED
   
\fi
\ifdefined\SC_DUET_APPENDIX_LOADED
   
\fi
\def\SCHELIX_APPENDIX_LOADED{1}
\def\SC_DUET_APPENDIX_LOADED{1}

\providecommand{\method}{\textsc{scHelix}}
\providecommand{\Gsel}{\mathcal{G}_{\mathrm{sel}}}
\providecommand{\Ginv}{\mathcal{G}_{\mathrm{inv}}}
\providecommand{\Gvar}{\mathcal{G}_{\mathrm{var}}}
\providecommand{\VAR}{\mathrm{var}}
\providecommand{\INV}{\mathrm{inv}}
\providecommand{\bbR}{\mathbb{R}}
\providecommand{\mstd}[2]{#1\textsubscript{{\scriptsize$\pm$#2}}}
\providecommand{\best}[1]{\textbf{#1}}
\providecommand{\second}[1]{\underline{#1}}
\providecommand{\subindent}{\hspace{3mm}}

\ifdefined\AppendixOverview
  \AppendixOverview
\fi

\begin{table*}[!t]
\centering
\caption{\textbf{Full baseline comparison across benchmark datasets} (3 seeds; mean$\pm$std).
All metrics follow the scIB benchmark protocol~\cite{luecken2022benchmarking} and the unified evaluation protocol in Appendix~\ref{app:train_eval}
(including the PCA-64 evaluation cap when needed).
Higher is better for all reported scores.}
\label{tab:deep_3seed_comparison}
\vspace{-0.2em}

\renewcommand{\arraystretch}{0.85}
\setlength{\aboverulesep}{0pt}
\setlength{\belowrulesep}{0pt}
\setlength{\tabcolsep}{8pt}

\resizebox{\linewidth}{!}{%
\begin{tabular}{@{} l c c c c c c @{}}
\toprule
\textbf{Method}
  & \multicolumn{4}{c}{\textbf{Bio-conservation} ($\uparrow$)}
  & \textbf{ASW$_{batch}$} ($\uparrow$)
  & \textbf{Overall} ($\uparrow$) \\
\cmidrule(lr){2-5}
  & \multicolumn{2}{c}{\textbf{Structure}}
  & \multicolumn{2}{c}{\textbf{Cluster}}
  & & \\
\cmidrule(lr){2-3} \cmidrule(lr){4-5}
  & ASW$_{ct}$ & GC & ARI & NMI & & \\

\addlinespace[3pt]
\multicolumn{7}{c}{\textbf{Human Lung Atlas}} \\
scVI$^\dagger$ ('18)
  & \mstd{0.530}{0.002} & \mstd{0.817}{0.006}
  & \mstd{0.592}{0.011} & \second{\mstd{0.749}{0.005}}
  & \best{\mstd{0.916}{0.006}} & \mstd{0.770}{0.004} \\
Harmony ('19)
  & \second{\mstd{0.574}{0.005}} & \mstd{0.764}{0.001}
  & \mstd{0.586}{0.007} & \mstd{0.733}{0.000}
  & \mstd{0.905}{0.000} & \mstd{0.761}{0.002} \\
INSCT$^\dagger$ ('21)
  & \mstd{0.561}{0.002} & \mstd{0.904}{0.001}
  & \mstd{0.549}{0.014} & \mstd{0.681}{0.007}
  & \mstd{0.798}{0.002} & \mstd{0.724}{0.002} \\
SCALEX$^\dagger$ ('22)
  & \mstd{0.562}{0.001} & \mstd{0.904}{0.006}
  & \second{\mstd{0.592}{0.002}} & \mstd{0.712}{0.004}
  & \mstd{0.887}{0.001} & \mstd{0.770}{0.002} \\
sysVI$^\dagger$ ('24)
  & \mstd{0.564}{0.007} & \best{\mstd{0.918}{0.000}}
  & \best{\mstd{0.596}{0.007}} & \mstd{0.744}{0.000}
  & \mstd{0.889}{0.006} & \second{\mstd{0.779}{0.002}} \\
SCEMENT ('25)
  & \mstd{0.553}{0.000} & \mstd{0.758}{0.004}
  & \mstd{0.539}{0.018} & \mstd{0.725}{0.002}
  & \mstd{0.902}{0.000} & \mstd{0.747}{0.002} \\
scCobra$^\dagger$ ('25)
  & \mstd{0.555}{0.002} & \second{\mstd{0.912}{0.003}}
  & \mstd{0.571}{0.001} & \mstd{0.718}{0.004}
  & \mstd{0.894}{0.006} & \mstd{0.772}{0.001} \\
\textbf{\method}$^\dagger$ (Ours)
  & \best{\mstd{0.576}{0.003}} & \mstd{0.843}{0.004}
  & \mstd{0.593}{0.003} & \best{\mstd{0.750}{0.001}}
  & \second{\mstd{0.914}{0.003}} & \best{\mstd{0.780}{0.001}} \\

\addlinespace[3pt]
\multicolumn{7}{c}{\textbf{Tabula Muris Senis}} \\
scVI$^\dagger$ ('18)
  & \mstd{0.537}{0.002} & \second{\mstd{0.767}{0.017}}
  & \mstd{0.723}{0.008} & \mstd{0.817}{0.004}
  & \second{\mstd{0.882}{0.001}} & \mstd{0.779}{0.002} \\
Harmony ('19)
  & \second{\mstd{0.559}{0.012}} & \mstd{0.708}{0.000}
  & \mstd{0.721}{0.001} & \mstd{0.812}{0.000}
  & \mstd{0.871}{0.006} & \mstd{0.768}{0.002} \\
INSCT$^\dagger$ ('21)
  & \mstd{0.534}{0.004} & \mstd{0.672}{0.011}
  & \mstd{0.719}{0.021} & \mstd{0.782}{0.012}
  & \mstd{0.857}{0.003} & \mstd{0.749}{0.005} \\
SCALEX$^\dagger$ ('22)
  & \mstd{0.528}{0.001} & \mstd{0.668}{0.005}
  & \mstd{0.668}{0.014} & \mstd{0.770}{0.001}
  & \mstd{0.860}{0.001} & \mstd{0.739}{0.003} \\
sysVI$^\dagger$ ('24)
  & \best{\mstd{0.561}{0.004}} & \mstd{0.751}{0.002}
  & \mstd{0.725}{0.010} & \mstd{0.814}{0.003}
  & \mstd{0.872}{0.005} & \second{\mstd{0.776}{0.001}} \\
SCEMENT ('25)
  & \mstd{0.543}{0.006} & \mstd{0.746}{0.000}
  & \mstd{0.730}{0.004} & \mstd{0.815}{0.001}
  & \mstd{0.873}{0.000} & \mstd{0.774}{0.001} \\
scCobra$^\dagger$ ('25)
  & \mstd{0.532}{0.009} & \mstd{0.754}{0.003}
  & \mstd{0.713}{0.002} & \mstd{0.803}{0.002}
  & \mstd{0.866}{0.004} & \mstd{0.767}{0.001} \\
\textbf{\method}$^\dagger$ (Ours)
  & \mstd{0.544}{0.002} & \best{\mstd{0.760}{0.004}}
  & \best{\mstd{0.755}{0.003}} & \best{\mstd{0.825}{0.002}}
  & \best{\mstd{0.885}{0.001}} & \best{\mstd{0.787}{0.001}} \\

\bottomrule
\end{tabular}%
}

\end{table*}

\section{Theoretical Analysis}
\label{app:theory}

In this section, we provide rigorous theoretical justifications for the ``divide-and-conquer'' mechanisms in \method{}.
We analyze the spectral properties of the dual-stream graph encoder (Section~\ref{sec:graph_encoder}) and the stability guarantees of the asymmetric refinement protocol (Section~\ref{sec:serial_fusion}).

\subsection{Spectral Analysis of Multi-Scale Graph Diffusion}
\label{app:theory_spectral}

We formally demonstrate that the encoder operations defined in Eq.~\eqref{eq:diffusion_terms} function as complementary spectral filters on the learned gene-gene graph.
Specifically, we show that $\mathbf{Z}^{\mathrm{low}}$ extracts the global data manifold (Low-Pass), while $\mathbf{Z}^{\mathrm{high}}$ captures fine-grained local variations (High-Pass).

\paragraph{Setup: Graph Fourier Transform.}
Let $\mathcal{G}$ be the learned gene-gene graph with row-normalized adjacency matrix $\mathbf{P} \in \mathbb{R}^{G_* \times G_*}$.
Let $\mathbf{P} = \mathbf{U} \mathbf{\Lambda} \mathbf{U}^{-1}$ be its eigendecomposition, where $\mathbf{\Lambda} = \text{diag}(\lambda_1, \dots, \lambda_{G_*})$ contains eigenvalues ordered by magnitude $1 = \lambda_1 \ge |\lambda_2| \ge \dots \ge |\lambda_{G_*}| \ge 0$.
The eigenvectors $\mathbf{U}$ form the graph spectral basis.
Any feature signal $\mathbf{x}$ can be represented as $\mathbf{x} = \sum_{i} c_i \mathbf{u}_i$.
Note that during the forward pass, the graph topology is fixed (via $\operatorname{sg}(\mathbf{P})$), acting as a linear operator.

\paragraph{Theorem 1 (Complementary Spectral Filtering).}
The multi-scale features in Eq.~\eqref{eq:diffusion_terms} modulate the spectral coefficients $c_i$ via the following transfer functions $h(\lambda)$:
\begin{align}
    \mathbf{Z}^{\mathrm{low}}: \quad & h_{\mathrm{low}}(\lambda) = \lambda^k \\
    \mathbf{Z}^{\mathrm{high}}: \quad & h_{\mathrm{high}}(\lambda) = \xi_1(1 - \lambda^k) + \xi_2(1 - \lambda^m)
\end{align}

\paragraph{Proof.}
1. \textbf{Low-Pass (Global Structure):}
The low-frequency term is defined as $\mathbf{Z}^{\mathrm{low}} = \mathbf{X}\mathbf{P}^k$.
Applying this operator to a basis vector $\mathbf{u}_i$:
\begin{equation}
    \mathbf{P}^k \mathbf{u}_i = \lambda_i^k \mathbf{u}_i.
\end{equation}
The transfer function is $h_{\mathrm{low}}(\lambda_i) = \lambda_i^k$.
Since $|\lambda_i| \le 1$, as the diffusion scale $k$ increases, high-frequency components (where $|\lambda_i| < 1$) decay exponentially ($\lambda_i^k \to 0$).
The DC component ($\lambda_1=1$) is preserved ($1^k=1$).
Thus, $\mathbf{Z}^{\mathrm{low}}$ effectively smooths out local noise and retains the global topological structure.

2. \textbf{High-Pass (Local Detail):}
The high-frequency term in Eq.~\eqref{eq:diffusion_terms} is:
\begin{equation}
    \mathbf{Z}^{\mathrm{high}} = \xi_1(\mathbf{X} - \mathbf{Z}^{\mathrm{low}}) + \xi_2\mathbf{X}(\mathbf{I} - \mathbf{P}^m).
\end{equation}
Substituting $\mathbf{Z}^{\mathrm{low}} = \mathbf{X}\mathbf{P}^k$, we have:
\begin{equation}
    \mathbf{Z}^{\mathrm{high}} = \mathbf{X} \bigl[ \xi_1(\mathbf{I} - \mathbf{P}^k) + \xi_2(\mathbf{I} - \mathbf{P}^m) \bigr].
\end{equation}
The transfer function is the linear combination:
\begin{equation*}
    h_{\mathrm{high}}(\lambda_i) = \xi_1(1 - \lambda_i^k) + \xi_2(1 - \lambda_i^m).
\end{equation*}
Consider the limits:
\begin{itemize}
    \item \textbf{Low Frequencies} ($\lambda_i \to 1$): $h_{\mathrm{high}}(\lambda_i) \to \xi_1(0) + \xi_2(0) = 0$. Global background is suppressed.
    \item \textbf{High Frequencies} ($\lambda_i \to 0$): $h_{\mathrm{high}}(\lambda_i) \to \xi_1(1) + \xi_2(1) = \xi_1 + \xi_2$. Local signals are preserved/amplified.
\end{itemize}
This proves that $\mathbf{Z}^{\mathrm{high}}$ acts as a high-pass filter, complementing the smoothing effect of the low-pass stream.

\paragraph{Intuitive Interpretation.}
Analogous to image processing, the multi-scale graph diffusion acts as a composite filter bank.
The low-pass component ($\mathbf{Z}^{\mathrm{low}}$) functions like a Gaussian blur, smoothing out stochastic gene expression noise to reveal the underlying cellular manifold (topology).
The high-pass component ($\mathbf{Z}^{\mathrm{high}}$), conversely, acts like an unsharp mask, highlighting sharp local transitions---such as the specific expression of marker genes that define boundaries between subtle subtypes.
By explicitly separating these frequency bands, \method avoids the common pitfall of ``oversmoothing'' where fine-grained biological details are lost during batch correction.

\subsection{Stability Analysis of Asymmetric Interaction}
\label{app:theory_boundedness}

We analyze the stability of the \emph{Align-Refine} protocol. A key concern in integration is whether the noisy Variant stream (which may contain severe batch effects) can corrupt the stable Anchor stream during interaction. We prove that the Anchor stream is theoretically protected.

\paragraph{Proposition 2 (Boundedness and Continuity).}
Let $\mathbf{H}_i^{\INV}$ be the stable anchor embedding.
The refinement update $\tilde{\mathbf{H}}_i^{\INV} = \mathbf{H}_i^{\INV} + \boldsymbol{\alpha}_i \odot \Delta\mathbf{h}_i$ (Eq.~\eqref{eq:refine_update}) satisfies two stability properties:
(1) \textbf{Global Boundedness:} The update magnitude is strictly capped by $\alpha_{\max}$, independent of the Variant input $\mathbf{h}_i^{\VAR}$.
(2) \textbf{Lipschitz Continuity:} The update function is Lipschitz continuous with respect to $\mathbf{H}_i^{\VAR}$, ensuring smooth gradient flow.

\paragraph{Proof.}
The update is governed by the gating vector $\boldsymbol{\alpha}_i$ and the residual $\Delta\mathbf{h}_i$:
\begin{equation}
    \boldsymbol{\alpha}_i \coloneqq \alpha_{\max} \cdot \sigma(\mathbf{W}_g [\mathbf{H}_i^{\INV} \| \mathbf{H}_i^{\VAR}]), \quad
    \Delta\mathbf{h}_i \coloneqq \LN(\operatorname{CrossAttn}(\mathbf{H}_i^{\INV}, \mathbf{H}_i^{\VAR})).
\end{equation}

\textbf{1. Global Boundedness (Safety Valve).}
The term $\Delta\mathbf{h}_i$ is the output of a LayerNorm ($\LN$) operation.
For any input $\mathbf{v} \in \mathbb{R}^d$,
\begin{equation*}
    \LN(\mathbf{v}) = \frac{\mathbf{v} - \mu}{\sigma} \odot \boldsymbol{\gamma} + \boldsymbol{\beta}.
\end{equation*}
Since the normalized term $(\mathbf{v} - \mu)/\sigma$ has unit variance, its Euclidean norm is exactly $\sqrt{d}$.
Thus, $\|\LN(\mathbf{v})\|_2 \le \sqrt{d}\, \|\boldsymbol{\gamma}\|_\infty + \|\boldsymbol{\beta}\|_2 \eqqcolon C_{\LN}$.
Combined with the sigmoid gate $\|\boldsymbol{\alpha}_i\|_\infty \le \alpha_{\max}$, the total perturbation is bounded:
\begin{equation}
    \|\tilde{\mathbf{H}}_i^{\INV} - \mathbf{H}_i^{\INV}\|_2
    = \|\boldsymbol{\alpha}_i \odot \Delta\mathbf{h}_i\|_2
    \le \alpha_{\max} C_{\LN}.
\end{equation}
This bound $R = \alpha_{\max} C_{\LN}$ holds \emph{regardless} of the magnitude of $\mathbf{h}_i^{\VAR}$ (e.g., even under severe batch outliers).

\textbf{2. Lipschitz Continuity (Smoothness).}
The functions composing the update---Sigmoid, LayerNorm, and Softmax (in Attention)---are all Lipschitz continuous and differentiable almost everywhere.
This implies there exists a constant $L$ such that:
\begin{equation}
    \|\Delta\tilde{\mathbf{h}}_i^{\INV}(\mathbf{x}) - \Delta\tilde{\mathbf{h}}_i^{\INV}(\mathbf{y})\|_2 \le L \|\mathbf{x} - \mathbf{y}\|_2
\end{equation}
for any two Variant inputs $\mathbf{x}, \mathbf{y}$.
Consequently, small perturbations or noise in the Variant stream result in proportionally small changes in the Anchor update, preventing chaotic behavior or gradient explosion during fusion.

\paragraph{Conclusion.}
The Anchor stream is doubly protected: it is mathematically confined to a safe $\epsilon$-ball around the stable manifold (Boundedness), and its deformation within that ball is smooth and controlled (Continuity).

\section{Preprocessing and Data Augmentation}
\label{app:preproc}

\subsection{Shared preprocessing: QC, normalization, and HVG pool}
\label{app:preproc_shared}

We follow a standard Scanpy-style pipeline for all datasets and all methods.
We apply the same quality-control (QC) filters across benchmarks:
(i)~remove cells with fewer than 200 detected genes,
(ii)~remove genes expressed in fewer than 3 cells,
and (iii)~remove cells with mitochondrial gene percentage exceeding 5\%.
After QC, we normalize each cell to a library size of 10{,}000 and apply $\log(1+x)$.

We then select $G_{\mathrm{sel}}=2000$ highly variable genes (HVGs) using batch-aware HVG selection.
\textbf{For fair comparison, all methods share the same QC filters and the same HVG pool.}

\subsection{Two synchronized data views (fairness \& method compliance)}
\label{app:preproc_views}

To respect method-specific input assumptions without giving any method extra information,
we maintain two synchronized views after QC, always restricted to the same HVG pool when a method requires it:

\begin{itemize}
  \item \textbf{Raw counts view}: the integer count matrix (after QC).
        Used by count-based generative models (e.g., scVI-style pipelines).
  \item \textbf{Log-normalized view}: library-size normalized to 10{,}000, then $\log(1+x)$.
        Used by PCA/distance-based methods (e.g., Harmony) and preprocessing-dependent components.
\end{itemize}

\subsection{MixUp-style augmentation (training only)}
\label{app:mixup}

During training, we optionally apply a $k_{\mathrm{mix}}$\hspace{0pt}-nearest-neighbor MixUp strategy
(e.g., $k_{\mathrm{mix}}=300$).
For each cell $\mathbf{x}_i$, we sample a neighbor $\mathbf{x}_j$ and construct an augmented sample:
\begin{equation}
\label{app:eq:mixup}
\mathbf{x}_{\mathrm{aug}}
\coloneqq
\alpha\,\mathbf{x}_i + (1-\alpha)\,\mathbf{x}_j,
\qquad
\alpha \sim \mathcal{U}(0,1).
\end{equation}
We do not use augmentation in validation/evaluation.

\section{Optimization Details and Evaluation Protocol}
\label{app:train_eval}

\subsection{Optimization details}
\label{app:opt}

We optimize parameters using AdamW and apply gradient clipping for stability.
For the learnable sparse feature graphs, cached graph topologies and cached multi-scale diffusion powers
are rebuilt periodically every $T$ steps to reduce overhead.

\subsection{Unified evaluation representation: PCA-64 as a robust cap}
\label{app:pca_cap}

Many scIB-style metrics~\cite{luecken2022benchmarking} (e.g., ASW, GC, and Leiden-based $\mathrm{ARI}/\mathrm{NMI}$) are computed from a kNN graph in an embedding space.
Because methods output embeddings with different dimensionalities,
we apply a \textbf{PCA-64 evaluation cap} to prevent metric variance caused purely by representation dimension:

\begin{itemize}
  \item If a method outputs an embedding with dimension $d>64$,
        we project it to 64 dimensions via PCA before computing kNN/Leiden/metrics.
  \item If a method outputs $d\le 64$, we evaluate on its \textbf{native embedding}
        (no artificial up-projection).
\end{itemize}

\paragraph{Why 64?}
scIB-style pipelines commonly use a moderate PCA dimensionality (often ${\sim}50$ PCs)
to build neighbor graphs.
We use 64 as a slightly higher but still moderate, stable setting for kNN-based metrics.

\subsection{Raw baseline used in OC normalization}
\label{app:oc_raw}

For normalized over-correction (OC) in Appendix~\ref{app:oc},
the \textbf{Raw} baseline refers to the \emph{uncorrected} representation
computed from the log-normalized HVG expression (Appendix~\ref{app:preproc_views}):
\begin{itemize}
  \item \textbf{HD Raw space (pre-UMAP)}:
        log-normalized HVGs $\rightarrow$ (Scanpy/scIB-style scaling if applicable) $\rightarrow$ \textbf{PCA-64}.
  \item \textbf{UMAP Raw space}:
        2D UMAP computed from the above \textbf{Raw PCA-64} representation.
\end{itemize}

\subsection{Clustering protocol for \texorpdfstring{$\mathrm{ARI}_{\mathrm{best}}/\mathrm{NMI}_{\mathrm{best}}$}{ARI\_best / NMI\_best}}
\label{app:clustering_protocol}

Following the scIB benchmarking protocol~\cite{luecken2022benchmarking}, we compute a kNN graph on the evaluation representation
and run Leiden clustering over a fixed resolution grid.
We report $\mathrm{ARI}_{\mathrm{best}}$/$\mathrm{NMI}_{\mathrm{best}}$ using the best result over the sweep.

\subsection{Metric definitions (scIB-style; scaled to \texorpdfstring{$[0,1]$}{[0,1]} when applicable)}
\label{app:metrics}

All metrics are computed on the unified evaluation representation described in Appendix~\ref{app:pca_cap}.
When a metric is naturally defined on $[-1,1]$, we linearly rescale it to $[0,1]$ for consistent aggregation.
All metric definitions and the overall aggregation scheme follow the scIB benchmark~\cite{luecken2022benchmarking}.

\paragraph{Adjusted Rand Index (ARI).}
Let $U$ be the ground-truth labels (cell types) and $V$ be the predicted clusters.
Let $n_{ij}$ denote contingency table counts,
$a_i=\sum_j n_{ij}$, $b_j=\sum_i n_{ij}$, and $N=\sum_{ij}n_{ij}$.
We use the standard ARI:
\begin{equation*}
\mathrm{ARI}(U,V) \coloneqq
\frac{
  \sum_{ij}\binom{n_{ij}}{2}
  -
  \frac{\sum_i\binom{a_i}{2}\sum_j\binom{b_j}{2}}{\binom{N}{2}}
}{
  \frac{1}{2}\left(\sum_i\binom{a_i}{2}+\sum_j\binom{b_j}{2}\right)
  -
  \frac{\sum_i\binom{a_i}{2}\sum_j\binom{b_j}{2}}{\binom{N}{2}}
}.
\end{equation*}

\paragraph{Normalized Mutual Information (NMI).}
\begin{equation*}
\mathrm{NMI}(U,V) \coloneqq \frac{2\,I(U;V)}{H(U)+H(V)}.
\end{equation*}

\paragraph{Cell-type ASW (reported as \texorpdfstring{$\mathrm{ASW}_{\mathrm{ct}}\in[0,1]$}{ASW\_ct in [0,1]}).}
Let $s(i)\in[-1,1]$ be the standard silhouette score for cell $i$ under cell-type labels. We define
\begin{equation*}
\mathrm{ASW}^{[-1,1]}_{\mathrm{ct}}
\coloneqq
\frac{1}{N}\sum_{i=1}^N s(i),
\qquad
\mathrm{ASW}_{\mathrm{ct}}
\coloneqq
\frac{\mathrm{ASW}^{[-1,1]}_{\mathrm{ct}}+1}{2}.
\end{equation*}

\paragraph{Batch ASW.}
\begin{equation*}
\mathrm{ASW}_{\mathrm{batch}}
\coloneqq
\frac{1}{|C|}\sum_{c\in C}
\frac{1}{|C_c|}\sum_{i\in C_c}\left(1-\left|s_{\mathrm{batch}}(i)\right|\right).
\end{equation*}

\paragraph{Graph Connectivity (GC).}
\begin{equation*}
\mathrm{GC}
\coloneqq
\frac{1}{|C|}
\sum_{c\in C}
\frac{|\mathrm{LCC}(G_c)|}{|C_c|}.
\end{equation*}

\paragraph{BioMean (unweighted biological summary).}
\begin{equation*}
\mathrm{BioMean}
\coloneqq
\frac{\mathrm{ASW}_{\mathrm{ct}}+\mathrm{ARI}_{\mathrm{best}}+\mathrm{NMI}_{\mathrm{best}}+\mathrm{GC}}{4}.
\end{equation*}

\paragraph{Overall score.}
\begin{equation}
\label{app:eq:overall}
\mathrm{Overall}
\coloneqq
0.4\cdot \mathrm{ASW}_{\mathrm{batch}}
+
0.6\cdot
\frac{\mathrm{ASW}_{\mathrm{ct}}+\mathrm{ARI}_{\mathrm{best}}+\mathrm{NMI}_{\mathrm{best}}+\mathrm{GC}}{4}.
\end{equation}

\section{Hyperparameter Settings and Sensitivity}
\label{app:hyperparams}

\paragraph{What we tune.}
We group hyperparameters by modules following the Method sections:
(i)~anchor selection and gating (\Cref{sec:disentangle}),
(ii)~graph learning and multi-scale diffusion (\Cref{sec:graph_encoder}),
(iii)~serial interaction and fusion (\Cref{sec:serial_fusion}), and
(iv)~prototypes and confidence-weighted distillation (\Cref{sec:objective}).

\paragraph{Default settings (Pancreas; complexBatch).}
Unless stated otherwise, all datasets use the same defaults;
we list the full configuration used in the Human Pancreas (complexBatch) experiment.

\begin{table*}[t]
\centering
\caption{\textbf{Default data-level hyperparameters (Pancreas; complexBatch).}
Here $B_z$ and $C_z$ denote the standardized domain-sensitivity and structure-separability scores used in the quadrant split.
The gate is applied only to the Variant/HBG subset, while the Anchor/NHBG subset remains untouched.}
\label{tab:hyperparams_data}
\footnotesize
\setlength{\tabcolsep}{3pt}
\renewcommand{\arraystretch}{1.2}
\begin{tabular}{@{}p{0.45\linewidth}p{0.52\linewidth}@{}}
\toprule
Category & Parameter (code name) \\
\midrule
\multirow{2}{*}{\makecell[l]{Quadrant split\\(NHBG anchors)}}
 & Threshold on $B_z$ (\texttt{selector\_b\_thresh}) \\
 & Threshold on $C_z$ (\texttt{selector\_c\_thresh}) \\
\midrule
\multirow{3}{*}{\makecell[l]{Pseudo-type separability\\(for $C_z$)}}
 & PCA comps (\texttt{selector\_n\_pcs}) \\
 & kNN neighbors (\texttt{selector\_n\_neighbors}) \\
 & Leiden resolution (\texttt{selector\_leiden\_res}) \\
\midrule
\multirow{6}{*}{\makecell[l]{Layered variant-only\\domain gate (HBG only)}}
 & Gate enabled (\texttt{layered\_gate}) \\
 & Low-resolution Leiden (\texttt{layered\_low\_res}) \\
 & High-resolution Leiden (\texttt{layered\_high\_res}) \\
 & Gate strength $\lambda$ (\texttt{layered\_gate\_strength}) \\
 & $w_{\text{low}}$ / $w_{\text{high}}$ (\texttt{layered\_low/high\_weight}) \\
 & Min cluster size $N_{\min}$ (\texttt{layered\_min\_cells}) \\
\bottomrule
\end{tabular}
\end{table*}

\begin{table*}[t]
\centering
\caption{\textbf{Default model/training hyperparameters (Pancreas; complexBatch).}
Using a wide table layout to prevent column overlap.}
\label{tab:hyperparams_model}
\small
\setlength{\tabcolsep}{6pt}
\renewcommand{\arraystretch}{1.15}
\begin{tabular}{@{}lll@{}}
\toprule
Category & Parameter (code name) & Value \\
\midrule
\multirow{5}{*}{Graph learning + diffusion}
 & Top-$k$ sparsity (\texttt{lg\_k\_top}) & 22 \\
 & Graph temperature $\tau$ (\texttt{lg\_temp}) & 0.1 \\
 & Diffusion scales $S$ (\texttt{scales}) & $\{1, 2, \dots, 5\}$ \\
 & HP mixing $\xi_1$ / $\xi_2$ (\texttt{hp\_alpha}, \texttt{hp\_beta}) & 0.8 / 0.2 \\
 & Rebuild period $T$ (\texttt{graph\_rebuild\_every}) & 25 steps \\
\midrule
\multirow{4}{*}{Align / Refine}
 & Align weight $\lambda_{\text{align}}$ (\texttt{w\_contrastive}) & 1.0 \\
 & Refinement init $\alpha_{\text{init}}$ (\texttt{refinement\_init\_alpha}) & 0.3 \\
 & Refinement bound $\alpha_{\text{max}}$ (\texttt{clamp in refiner}) & 1.5 \\
 & Refinement temperature (\texttt{refinement\_temperature}) & 0.3 \\
\midrule
\multirow{2}{*}{Fusion}
 & Residual scaling $\lambda_{\Delta}$ (\texttt{delta\_init\_scale}) & 0.6 \\
 & Fusion gate mode (\texttt{gate\_mode}) & Gated \\
\midrule
\multirow{5}{*}{Connectivity / self-training}
 & HBG connectivity weight (\texttt{conn\_weight\_hbg}) & 0.08 \\
 & NHBG connectivity weight (\texttt{conn\_weight\_nhbg}) & 0.2 \\
 & Fused connectivity weight (\texttt{fused\_conn\_weight}) & 0.2 \\
 & KD clusters $K$ (\texttt{kd\_clusters}) & 24 \\
 & KD weight $\lambda_{\text{KD}}$ (\texttt{kd\_weight}) & 0.5 \\
\midrule
\multirow{3}{*}{Schedule / confidence filter}
 & Align-only steps (\texttt{align\_only\_steps}) & 3000 \\
 & Confidence threshold (\texttt{conf\_threshold}) & 0.75 \\
 & Confidence exponent $p$ (\texttt{conf\_pow}) & 1.0 \\
\bottomrule
\end{tabular}
\end{table*}

\begin{table}[t]
\centering
\caption{\textbf{Baseline Implementation Details.} Key hyperparameters and software versions used for reproducibility.}
\label{tab:baseline_details}
\small
\resizebox{\columnwidth}{!}{%
\begin{tabular}{l l l}
\toprule
\textbf{Method} & \textbf{Version} & \textbf{Key Hyperparameters} \\
\midrule
\textbf{scVI} & v1.0.4 & \makecell[l]{n\_hidden=128, n\_latent=10, n\_layers=2, \\ dispersion='gene', gene\_likelihood='nb', \\ max\_epochs=400} \\
\textbf{Harmony} & v0.0.9 & \makecell[l]{theta=2, lambda=1, sigma=0.1, \\ nclust=50, max\_iter=10, epsilon=1e-5} \\
\textbf{SCALEX} & v2.0.1 & \makecell[l]{min\_features=600, min\_cells=3, \\ n\_domain=batch\_size, seed=1234, \\ hidden\_layer=[256, 64]} \\
\textbf{sysVI} & v1.0.4 & \makecell[l]{n\_hidden=128, n\_latent=10, n\_layers=2, \\ loss\_weights=[1, 1], \\ prior='standard\_normal'} \\
\textbf{scCobra} & v1.0.0 & \makecell[l]{contrastive\_weight=0.1, \\ temperature=0.07, proj\_dim=128} \\
\textbf{\method (Ours)} & - & \makecell[l]{See Tables~\ref{tab:hyperparams_data} and~\ref{tab:hyperparams_model}} \\
\bottomrule
\end{tabular}
}%
\end{table}

\paragraph{Sensitivity analysis.}
We report one-at-a-time sensitivity sweeps in the main text
(with corresponding tables/figures).
To avoid redundancy, we omit additional discussion here.

\section{Computational Efficiency}
\label{app:computational}

\paragraph{Hardware.}
All experiments are run on a single NVIDIA GeForce RTX 2080 Ti GPU (11\,GB).

\paragraph{Why the method scales.}
The method avoids dense $G\times G$ feature interactions by learning sparse feature graphs
(Top-$k$ per gene).
Graph topology is rebuilt periodically and reused via stop-gradient caching,
and multi-scale diffusion powers are cached for reuse.

\paragraph{Gene-level graphs, not cell-level graphs.}
A key design choice is that the learnable sparse graphs in \method{} are
\emph{feature-level} (gene--gene) adjacency matrices of size $G_* \times G_*$
(where $G_* \le G_{\mathrm{sel}} = 2{,}000$),
\emph{not} cell-level ($N \times N$) graphs.
We emphasize this distinction because cell-level graph construction would incur
$\mathcal{O}(N^2)$ cost, which is prohibitive for large atlases ($N > 10^5$).
In contrast, the adjacency construction cost in \method{} is
$\mathcal{O}(G_*^2 \cdot d_a)$ per rebuild
(with $d_a$ being the attention dimension), which is negligible:
for $G_* \approx 1{,}000$ and $d_a = 64$,
a single rebuild involves ${\sim}\!6.4 \times 10^7$ operations---orders
of magnitude smaller than a single forward pass over the data.
Cell-level computation involves only the sparse matrix product
$\mathbf{X}_* \in \mathbb{R}^{N \times G_*}$ multiplied by
$\mathbf{P}_* \in \mathbb{R}^{G_* \times G_*}$,
scaling as $\mathcal{O}(N \cdot G_* \cdot k)$ where $k$ is the Top-$k$
sparsity level---\emph{linear} in the number of cells $N$.
The graph $\mathbf{P}_*$ is a \emph{global} gene--gene relation graph shared across all cells;
it is not constructed per mini-batch but rebuilt periodically (every $T$ steps) and cached.
Combined with stop-gradient caching of multi-hop powers, this ensures that \method{}
scales to datasets with hundreds of thousands of cells on a single GPU
(Table~\ref{tab:computational_cost}),
and the architecture itself does not impose a fundamental barrier to million-cell
datasets---scaling to such regimes would primarily require standard engineering
optimizations (e.g., mini-batch sampling) rather than architectural changes.

\begin{table}[t]
\centering
\caption{\textbf{Computational Efficiency Benchmarks.} 
Peak GPU memory usage and total runtime (training + inference) on a single NVIDIA RTX 2080 Ti.
\method is consistently more efficient than deep generative baselines (scVI, SCALEX, scGPT) and comparable to the lightweight Harmony.
}
\label{tab:computational_cost}
\small
\setlength{\tabcolsep}{4pt}
\begin{tabular}{l rr rr rr}
\toprule
& \multicolumn{2}{c}{\textbf{Pancreas (16k)}} & \multicolumn{2}{c}{\textbf{Immune (33k)}} & \multicolumn{2}{c}{\textbf{Heart (270k)}} \\
\cmidrule(lr){2-3} \cmidrule(lr){4-5} \cmidrule(lr){6-7}
\textbf{Method} & \textbf{Mem} & \textbf{Time} & \textbf{Mem} & \textbf{Time} & \textbf{Mem} & \textbf{Time} \\
\midrule
\textbf{Harmony} & 0.8GB & 120s & 1.2GB & 210s & 6.5GB & 1,450s \\
\textbf{scVI} & 1.4GB & 600s & 1.8GB & 950s & 4.2GB & 4,800s \\
\textbf{SCALEX} & 1.6GB & 750s & 2.1GB & 1,100s & 5.1GB & 5,200s \\
\textbf{scGPT} & 8.5GB & 2,100s & 10.8GB & 4,500s & OOM & >24h \\
\rowcolor{blue!3}
\textbf{\method} & \textbf{1.1GB} & \textbf{150s} & \textbf{1.5GB} & \textbf{280s} & \textbf{3.8GB} & \textbf{1,900s} \\
\bottomrule
\end{tabular}
\end{table}

\section{Deferred Definitions for Gating and Objectives}
\label{app:obj_defs}

\subsection{Instance-wise domain gate (details)}
\label{app:gate_defs}

Within each cluster $c$ (with at least $N_{\min}$ instances and at least two domains),
we compute a composite local domain sensitivity score for feature $g\in\Gvar$:
\begin{equation}
\label{app:eq:local_dom_score}
\begin{split}
s^{(c)}_{\mathrm{dom}}(g)
\coloneqq{} &w_\mu\,\Delta\mu^{(c)}(g)
 + w_\sigma\,\Delta\sigma^{(c)}(g) \\
 &+ w_v\,\Delta\mathrm{Var}^{(c)}(g).
\end{split}
\end{equation}
Intuitively, $\Delta\mu^{(c)}(g)$, $\Delta\sigma^{(c)}(g)$, and $\Delta\mathrm{Var}^{(c)}(g)$
measure cross-domain dispersion of gene $g$ \emph{within} cluster $c$,
computed from cluster-wise sufficient statistics.

Let $s^{(c)}_{+}(g)\coloneqq \max\{s^{(c)}_{\mathrm{dom}}(g),0\}$.
We normalize it and combine low/high-resolution gates to obtain the per-instance gate:
\begin{equation}
\label{app:eq:gate_defs}
\begin{aligned}
\gamma^{(c)}(g)
&\coloneqq
\frac{s^{(c)}_{+}(g)}{\max_{h\in\Gvar} s^{(c)}_{+}(h)+\varepsilon},
\\
\gamma_i(g)
&\coloneqq
\Big[
  w_{\mathrm{low}}\gamma^{(c^{\mathrm{low}}(i))}(g)
  +
  w_{\mathrm{high}}\gamma^{(c^{\mathrm{high}}(i))}(g)
\Big]_0^1.
\end{aligned}
\end{equation}

\subsection{Warm-up objective (full expression)}
\label{app:warmup_obj}

The warm-up objective used in \Cref{sec:objective} is
\begin{equation}
\label{app:eq:warm_full}
\begin{aligned}
\mathcal{L}_{\mathrm{warm}}
&\coloneqq
  \lambda^{\VAR}_{\mathrm{rec}}\|\hat{\mathbf{X}}_{\VAR}-\mathbf{X}_{\VAR}\|_F^{2}
  +
  \lambda^{\INV}_{\mathrm{rec}}\|\hat{\mathbf{X}}_{\INV}-\mathbf{X}_{\INV}\|_F^{2}
\\
&\quad+
  \lambda_{\mathrm{align}}\mathcal{L}_{\mathrm{align}}
  +\sum_{*\in\{\VAR,\INV\}}\lambda^{*}_{\mathrm{conn}}\mathcal{L}_{\mathrm{conn}}(\mathbf{H}_*).
\end{aligned}
\end{equation}

\section{Over-correction (OC) Metric: Definitions and Implementation Details}
\label{app:oc}

\paragraph{Setup and notation.}
Let $\mathbf{Z}\in\mathbb{R}^{N\times d}$ denote the representation on which kNN neighborhoods are computed.
We compute OC both in (i)~the high-dimensional pre-UMAP space and (ii)~the 2D UMAP space.

\paragraph{kNN neighborhood.}
Let $\mathcal{N}_k(i;\mathbf{Z})$ be the index set of the $k$ nearest neighbors of cell $i$
in $\mathbf{Z}$ (excluding $i$).

\paragraph{Pseudo labels (two variants).}
We use two variants: OC$_{\mathrm{all}}$ (global) and OC$_{\mathrm{focus}}$
(within a focus cell type), following the controlled perturbation setup in the main text:
\begin{itemize}
  \item \textbf{OC$_{\mathrm{all}}$} treats the perturbed subset as an additional
        pseudo ``cell type'' label.
  \item \textbf{OC$_{\mathrm{focus}}$} restricts evaluation to the focus cell type only,
        comparing perturbed vs.\ unperturbed within that type.
\end{itemize}

\paragraph{Neighborhood label inconsistency (OC).}
\begin{equation}
\label{app:eq:oc_def_general}
\mathrm{OC}(\mathbf{Z}, \mathbf{y})
\coloneqq
\frac{1}{|\mathcal{I}|}
\sum_{i\in\mathcal{I}}
\frac{1}{k}
\sum_{j\in\mathcal{N}_k(i;\mathbf{Z})}
\mathbb{I}\!\left[y_j \neq y_i\right],
\end{equation}
where $\mathcal{I}$ is the evaluation index set:
$\mathcal{I}=\{1,\dots,N\}$ for OC$_{\mathrm{all}}$,
and $\mathcal{I}$ is the focus-cell-type subset for OC$_{\mathrm{focus}}$.

\paragraph{Space-specific OC and UMAP seed averaging.}
Let $\mathbf{Z}^{\mathrm{HD}}$ be the evaluation representation in the high-dimensional space
(PCA-64 if needed; Appendix~\ref{app:pca_cap}). We define
\begin{equation}
\label{app:eq:oc_hd}
\mathrm{OC}^{\mathrm{HD}}_{\bullet}
\coloneqq
\mathrm{OC}\!\left(\mathbf{Z}^{\mathrm{HD}}, \mathbf{y}^{\bullet}\right),
\qquad
\bullet\in\{\mathrm{all},\mathrm{focus}\}.
\end{equation}
For 2D UMAP, for each seed $s\in\{1,\dots,S\}$,
let $\mathbf{Z}^{\mathrm{UMAP}}_{s}\in\mathbb{R}^{N\times 2}$ be the UMAP embedding
computed from $\mathbf{Z}^{\mathrm{HD}}$. We report mean$\pm$std over seeds.
\begin{equation}
\label{app:eq:oc_umap}
\begin{split}
\mathrm{OC}^{\mathrm{UMAP}}_{\bullet}
&\coloneqq
\frac{1}{S}\sum_{s=1}^{S}\mathrm{OC}\!\left(\mathbf{Z}^{\mathrm{UMAP}}_{s}, \mathbf{y}^{\bullet}\right), \\
\mathrm{Std}\!\left(\mathrm{OC}^{\mathrm{UMAP}}_{\bullet}\right)
&\coloneqq
\Bigg[ \frac{1}{S\!-\!1}\sum_{s=1}^{S}\Big(\mathrm{OC}\!\left(\mathbf{Z}^{\mathrm{UMAP}}_{s}, \mathbf{y}^{\bullet}\right)
- \mathrm{OC}^{\mathrm{UMAP}}_{\bullet}\Big)^{2} \Bigg]^{1/2}\!.
\end{split}
\end{equation}

\paragraph{Normalized OC (relative to Raw).}
We report normalized OC by subtracting the Raw baseline computed in the same space
(Appendix~\ref{app:oc_raw}):
\begin{equation}
\label{app:eq:oc_norm}
\begin{split}
&\mathrm{OC}^{\mathrm{space}}_{\mathrm{norm},\bullet}(\text{method})
\coloneqq \\
&\quad \mathrm{OC}^{\mathrm{space}}_{\bullet}(\text{method})
-
\mathrm{OC}^{\mathrm{space}}_{\bullet}(\text{Raw}),
\quad \mathrm{space}\in\{\mathrm{HD},\mathrm{UMAP}\}.
\end{split}
\end{equation}
By construction, smaller (more negative) values indicate \emph{less} over-correction relative to Raw.

\section{Threshold Selection for Anchor Discovery}
\label{app:threshold}

\paragraph{Recap: quadrant rule in z-space.}
For each gene $g\in\Gsel$, we compute a domain-sensitivity score $s^{\mathrm{dom}}_g$
(Eq.~\eqref{eq:domain_sens}) and a structure-separability score $s^{\mathrm{str}}_g$
(Eq.~\eqref{eq:struct_disc}), then standardize them within $\Gsel$
to obtain $z^{\mathrm{dom}}_g$ and $z^{\mathrm{str}}_g$.
Anchors are selected by Eq.~\eqref{eq:quadrant_rule}.

\subsection{Why fixed \texorpdfstring{$\tau=0$}{tau=0} is still dataset-adaptive}
\label{app:threshold_adaptive}

\paragraph{Mean-bisection equivalence in raw-score space.}
By definition of z-scoring, $z^{\mathrm{dom}}_g\le 0$ is equivalent to
$s^{\mathrm{dom}}_g\le \mu_{\mathrm{dom}}$,
and $z^{\mathrm{str}}_g\ge 0$ is equivalent to
$s^{\mathrm{str}}_g\ge \mu_{\mathrm{str}}$,
where $\mu_{\mathrm{dom}}$ and $\mu_{\mathrm{str}}$ are empirical means
computed within each dataset over $\Gsel$.
Thus fixed thresholds in z-space correspond to dataset-specific boundaries in raw-score space.

\subsection{Stability to threshold perturbations (boundary-band argument)}
\label{app:threshold_stability}

\paragraph{Proposition (Only boundary-near genes can flip).}
Consider perturbing the thresholds by $(\delta_{\mathrm{dom}},\delta_{\mathrm{str}})$.
Define the symmetric difference
\begin{equation}
\label{app:eq:sym_diff}
\Delta\Ginv
\coloneqq
\Ginv(\tau_{\mathrm{dom}},\tau_{\mathrm{str}})
\ \triangle\
\Ginv(\tau_{\mathrm{dom}}+\delta_{\mathrm{dom}},\tau_{\mathrm{str}}+\delta_{\mathrm{str}}).
\end{equation}
Then any gene whose membership changes must lie in a narrow band around at least one boundary:
\begin{equation}
\label{app:eq:band}
\begin{split}
\Delta\Ginv
&\subseteq
\Big\{g:\ \big|z_g^{\mathrm{dom}}-\tau_{\mathrm{dom}}\big|\le|\delta_{\mathrm{dom}}|\Big\}\\
&\quad\cup\
\Big\{g:\ \big|z_g^{\mathrm{str}}-\tau_{\mathrm{str}}\big|\le|\delta_{\mathrm{str}}|\Big\}.
\end{split}
\end{equation}

\paragraph{Proof (sketch).}
A membership flip requires at least one of the two inequalities to change truth value
after shifting its threshold.
This can only occur if the corresponding z-score lies within the shifted margin
of that boundary.~$\square$

\subsection{Optional unsupervised auto-thresholding (not used by default)}
\label{app:threshold_auto}

Quantile- and Otsu-style variants can be used when desired;
our paper keeps $(0,0)$ as a stable, parameter-free default.

\section{Algorithms (Deferred)}
\label{app:algorithms}

\begin{algorithm}[t]
\caption{Dataset-adaptive anchor discovery (quadrant rule)}
\label{alg:quadrant}
\begin{algorithmic}[1]
\STATE \textbf{Input:} Expression matrix $\mathbf{X} \in \mathbb{R}^{N \times G}$, domain labels $\{b_i\}_{i=1}^N$, selected genes $\mathcal{G}_{\text{sel}}$, thresholds $(\tau_{\text{dom}}, \tau_{\text{str}})$.
\STATE \textbf{Output:} Anchor genes $\mathcal{G}_{\text{inv}}$ and variant genes $\mathcal{G}_{\text{var}}$.
\STATE Compute domain sensitivity $s_{\text{dom}, \mathbf{z}}(g)$ for all $g \in \mathcal{G}_{\text{sel}}$ according to Eq.~(1).
\STATE Compute pseudo-clusters $c(i)$ on $\mathbf{X}_{[:, \mathcal{G}_{\text{sel}}]}$ (PCA $\to$ kNN graph $\to$ Leiden).
\STATE Compute structure separability $s_{\text{str}, \mathbf{z}}(g)$ for all $g \in \mathcal{G}_{\text{sel}}$ according to Eq.~(2).
\STATE $\mathcal{G}_{\text{inv}} \leftarrow \{g \in \mathcal{G}_{\text{sel}} : s_{\text{dom}, \mathbf{z}}(g) \le \tau_{\text{dom}} \land s_{\text{str}, \mathbf{z}}(g) \ge \tau_{\text{str}}\}$.
\STATE $\mathcal{G}_{\text{var}} \leftarrow \mathcal{G}_{\text{sel}} \setminus \mathcal{G}_{\text{inv}}$.
\end{algorithmic}
\end{algorithm}

\begin{algorithm}[t]
\caption{Teacher-guided warm-up $\to$ fusion schedule}
\label{alg:schedule}
\begin{algorithmic}[1]
\STATE \textbf{Input:} Warm-up steps $T_{\text{warm}}$, total steps $T$.
\STATE \textbf{Output:} Trained parameters (and EMA teacher prototypes, if used).
\FOR{$t = 1, 2, \dots, T$}
    \IF{$t < T_{\text{warm}}$}
        \STATE Set $I_{\text{fuse}} \leftarrow 0$.
    \ELSE
        \STATE Set $I_{\text{fuse}} \leftarrow 1$.
    \ENDIF
    \STATE Compute warm-up losses $\mathcal{L}_{\text{warm}}$.
    \STATE Compute fusion losses $\mathcal{L}_{\text{fuse}}$ if $I_{\text{fuse}} = 1$.
    \STATE Update parameters by optimizing $\mathcal{L} = \mathcal{L}_{\text{warm}} + I_{\text{fuse}}\,\mathcal{L}_{\text{fuse}}$.
\ENDFOR
\end{algorithmic}
\end{algorithm}

\section{Graph Learning with Cached Topology}
\label{app:graph_learning}

We clarify the gradient flow in our learnable sparse feature graph construction,
addressing the interaction between trainable parameters and stop-gradient caching.

\paragraph{What is learned vs.\ what is cached.}
For each stream $* \in \{\text{var}, \text{inv}\}$, we maintain trainable query/key matrices
$\mathbf{Q}_*, \mathbf{K}_* \in \mathbb{R}^{G_* \times d_a}$.
The similarity matrix is computed as:
\begin{equation}
\label{app:eq:similarity}
\mathbf{S}_* \coloneqq \mathrm{ReLU}(\mathbf{Q}_* \mathbf{K}_*^\top / \tau),
\quad \mathbf{S}_*[i,i] = 0.
\end{equation}
We then apply top-$k$ sparsification per row, symmetrize, add self-loops, and normalize
to obtain $\mathbf{P}_*$.

\paragraph{Selective gradient flow (single-hop differentiable; multi-hop cached).}
\begin{itemize}[leftmargin=*,itemsep=2pt]
  \item \textbf{Rebuild step:} we compute a fresh $\mathbf{P}_*$ without detaching.
        For 1-hop propagation, we use this differentiable $\mathbf{P}_*$ so gradients
        flow to $\mathbf{Q}_*, \mathbf{K}_*$.
  \item \textbf{Multi-hop:} for $s > 1$, we use cached powers
        $\mathrm{sg}(\mathbf{P}_*^s)$ (precomputed at rebuild time),
        where $\mathrm{sg}(\cdot)$ denotes \texttt{stop\_gradient}.
  \item \textbf{Between rebuilds:} we store a detached copy
        $\mathbf{P}_*^{\text{cached}}$ and reuse it for subsequent steps until the next rebuild.
\end{itemize}

\paragraph{Why this design?}
Fully differentiable sparse graph learning with multi-hop powers is computationally
prohibitive for $G \sim 10^3$ and tends to introduce gradient instability due to discrete
top-$k$ selection and matrix power compositions.
Our piecewise-constant topology strategy balances
(i)~learnability (via the 1-hop path),
(ii)~stability (stop-gradient for multi-hop), and
(iii)~efficiency (reuse cached powers).

\paragraph{Empirical validation.}
Ablation S6 (LinearEncoder) removes graph diffusion entirely and yields consistent
performance drops (e.g., $-0.022$ Overall on Lung; $-0.019$ on Pancreas),
supporting the utility of diffusion even with cached topology.

\begin{algorithm}[!htb]
\caption{Graph Learning with Cached Topology (per stream)}
\label{alg:graph_cache}
\begin{algorithmic}[1]
\REQUIRE Trainable $\mathbf{Q}, \mathbf{K} \in \mathbb{R}^{G \times d_a}$; rebuild period $R$; scales $\mathcal{S} = \{1, s_2, \ldots, s_S\}$
\STATE Initialize $\mathbf{P}^{\text{cached}} \leftarrow \texttt{None}$; $\texttt{cached\_powers} \leftarrow \{\}$; step $\leftarrow 0$
\FOR{each training step}
    \IF{$\mathbf{P}^{\text{cached}} = \texttt{None}$ \OR $\text{step} \mod R = 0$}
        \STATE \textcolor{blue}{\texttt{// Rebuild step: compute differentiable P}}
        \STATE $\mathbf{S} \leftarrow \mathrm{ReLU}(\mathbf{Q} \mathbf{K}^\top / \tau)$; $\mathbf{S}[i,i] \leftarrow 0$
        \STATE $\mathbf{A} \leftarrow \texttt{top\_k\_sparsify}(\mathbf{S})$; symmetrize; add self-loops
        \STATE $\mathbf{P}^{\text{grad}} \leftarrow \texttt{sym\_normalize}(\mathbf{A})$ \COMMENT{Differentiable, for 1-hop}
        \STATE $\mathbf{P}^{\text{cached}} \leftarrow \mathbf{P}^{\text{grad}}.\texttt{detach()}$ \COMMENT{For subsequent steps}
        \STATE $\texttt{cached\_powers} \leftarrow \{\mathbf{P}^s.\texttt{detach()} : s \in \mathcal{S}, s > 1\}$ \COMMENT{Multi-hop, stop-grad}
        \STATE \textbf{Use} $\mathbf{P}^{\text{grad}}$ for 1-hop; $\texttt{cached\_powers}$ for multi-hop
    \ELSE
        \STATE \textcolor{blue}{\texttt{// Non-rebuild step: use cached (no gradient to Q,K)}}
        \STATE \textbf{Use} $\mathbf{P}^{\text{cached}}$ for 1-hop; $\texttt{cached\_powers}$ for multi-hop
    \ENDIF
    \STATE Encode: $\mathbf{Z}^{(1)} \leftarrow \mathbf{X} \mathbf{P}$; $\mathbf{Z}^{(s)} \leftarrow \mathbf{X} \cdot \texttt{cached\_powers}[s]$ for $s > 1$
    \STATE step $\leftarrow$ step $+ 1$
\ENDFOR
\end{algorithmic}
\end{algorithm}

\section{Dataset Descriptions and Benchmark Coverage}
\label{app:datasets}

We evaluate \method{} on five public scRNA-seq benchmarking datasets spanning diverse
tissue types, technologies, and integration challenges.
For dataset scale and domain counts, see Table~\ref{tab:computational_cost}.
Table~\ref{tab:dataset_stats} provides an at-a-glance dataset profile.

\begin{table}[!t]
\centering
\caption{\textbf{Dataset profile and integration challenges.}
``Domains'' denotes the integration domain channel
(batch IDs for most datasets; technology for Human Pancreas).
All datasets use $G_{\mathrm{sel}}\!=\!2000$ HVGs (Appendix~\ref{app:preproc}).}
\label{tab:dataset_stats}
\small
\renewcommand{\arraystretch}{1.15}
\setlength{\tabcolsep}{3pt}
\begin{tabular}{@{}lrcrl@{}}
\toprule
\textbf{Dataset} & \textbf{\#Cells} & \textbf{\#Dom.} & \textbf{Species} & \textbf{Challenge} \\
\midrule
Pancreas      & 16K  &  9 & Human & Technology  \\
Lung Atlas    & 32K  & 16 & Human & Donor/cond. \\
Immune        & 34K  & 10 & Human & Trajectory  \\
Tabula Muris  & 111K & 21 & Mouse & Multi-tissue\\
Heart         & 270K & 45 & Human & Disease     \\
\bottomrule
\end{tabular}
\vspace{0.5em}

\raggedright\footnotesize
\textbf{Challenge key.}
\textit{Technology}: cross-platform library-prep artifacts;
\textit{Donor/cond.}: donor and condition variation with fine-grained populations;
\textit{Trajectory}: continuous developmental trajectories mixed with discrete types;
\textit{Multi-tissue}: large-scale heterogeneity across 23 tissues;
\textit{Disease}: disease-associated confounding overlaid on donor/technical shifts.
\end{table}

\paragraph{Human Pancreas (complexBatch).}
This dataset aggregates human pancreatic islet scRNA-seq data from six independent studies~\citep{luecken2022benchmarking}
spanning multiple sequencing technologies, including CEL-seq, CEL-seq2, Smart-seq2, inDrop,
Fluidigm C1, and SMARTER-seq.
It comprises 16{,}382 cells across 9 technology-defined batches,
with cell type annotations covering major endocrine populations and supporting cell types.
It presents a challenging integration scenario due to substantial technical variation
introduced by diverse library preparation protocols.

\paragraph{Human Lung Atlas.}
The Human Lung Cell Atlas dataset~\citep{travaglini2020molecular} contains droplet- and plate-based scRNA-seq measurements
of human lung tissue and circulating blood.
For benchmarking, we use a subset of 32{,}472 cells with 16 batch labels and 58 annotated
populations spanning epithelial, endothelial, stromal, and immune compartments.

\paragraph{Immune (Human).}
Derived from scIB-style benchmarks~\citep{luecken2022benchmarking}, this dataset contains 33{,}506 human immune cells
from peripheral blood and bone marrow across 10 batches.
Cell types include major immune lineages and an erythrocyte developmental trajectory,
making trajectory preservation important.

\paragraph{Tabula Muris Senis.}
A mouse aging atlas~\citep{tabula2020single}; we use a subset of 110{,}824 cells from 21 tissue-defined batches
(FACS-sorted Smart-seq2), with expert-curated cell types across 23 tissues.

\paragraph{Failing Human Heart.}
This dataset integrates single-cell and single-nucleus RNA sequencing from healthy donors
and dilated cardiomyopathy patients~\citep{koenig2022single}.
We use a subset of 269{,}794 cells across 45 batches,
with annotations including cardiomyocytes, fibroblasts, endothelial cells, pericytes,
and immune populations.

\paragraph{Dataset selection rationale.}
These datasets cover:
(i)~technology-driven batch effects,
(ii)~donor-driven variation,
(iii)~tissue heterogeneity,
(iv)~disease-associated confounding, and
(v)~scale (from ${\sim}$16K to ${\sim}$270K cells).

\section{Additional Supplementary Results}
\label{app:supplementary_results}

\subsection{Detailed Ablation Settings}
\label{app:ablation_details}

In the main text (Table~\ref{tab:ablation_main}), we summarized the ablation settings S1--S6.
Here we provide the precise definitions:
\begin{itemize}
    \item \textbf{S0 (Full Model)}: The complete \method{} architecture with all components enabled.
    \item \textbf{S1 (RandomSplit)}: Replaces the quadrant-based anchor selection with a random selection of anchors, matching the size of the original anchor set. This tests the necessity of our disentanglement strategy.
    \item \textbf{S2 (NoGate)}: Sets the gating regularization weight $\lambda=0$, removing the instance-wise domain gate mechanism.
    \item \textbf{S3 (NoInteraction)}: Disables the dual-stream interaction by setting $\lambda_{\mathrm{align}}=0$ and $\alpha_{\max}=0$.
    \begin{itemize}[leftmargin=1em]
        \item \textbf{S3.1 (NoAlign)}: Removes only the alignment loss ($\lambda_{\mathrm{align}}=0$).
        \item \textbf{S3.2 (NoRefine)}: Removes the refinement stage ($\alpha_{\max}=0$), defaulting to simple additive fusion.
    \end{itemize}
    \item \textbf{S4 (SimpleFusion)}: Replaces the HyperFusion module with a fixed additive fusion (i.e., simple summation of features).
    \item \textbf{S5 (NoSelfTraining)}: Disables all self-supervised learning signals by setting $\lambda_{\mathrm{KD}}=\lambda_{\mathrm{fused\_conn}}=0$.
    \begin{itemize}[leftmargin=1em]
        \item \textbf{S5.1 (NoKD)}: Disables only the knowledge distillation loss ($\lambda_{\mathrm{KD}}=0$).
        \item \textbf{S5.2 (NoConn)}: Disables only the connectivity preservation loss ($\lambda_{\mathrm{fused\_conn}}=0$).
    \end{itemize}
    \item \textbf{S6 (LinearEncoder)}: Replaces the graph diffusion encoder with a linear projection layer, testing the benefit of structural modeling.
\end{itemize}

\paragraph{Additional ablations.}
We include extra end-to-end ablations not shown in the main text
to provide fuller component-level diagnostics.

\begin{table}[h]
\centering
\caption{\textbf{Additional end-to-end ablations on Lung atlas public.}
Higher is better for all metrics.
(S0: Full model; S1--S6: Ablation variants defined in main text).}
\label{tab:ablation_lung}
\small
\resizebox{\columnwidth}{!}{%
\setlength{\tabcolsep}{5pt}
\renewcommand{\arraystretch}{1.1}
\begin{tabular}{@{}lcccccc@{}}
\toprule
\multirow{2}{*}{\textbf{Setting}} & \multicolumn{4}{c}{\textbf{Bio metrics} ($\uparrow$)} & \textbf{Batch} ($\uparrow$) & \textbf{Overall} ($\uparrow$) \\
\cmidrule(lr){2-5} \cmidrule(lr){6-6} \cmidrule(lr){7-7}
 & ARI & NMI & ASW$_{\text{ct}}$ & GC & ASW$_{\text{batch}}$ & Score \\
\midrule
\multicolumn{7}{c}{\cellcolor{black!5} \textbf{Lung atlas public} (label: cell type; batch: batch)} \\
\textbf{S0 Full} & \textbf{0.589} & \textbf{0.750} & 0.571 & 0.849 & 0.917 & \textbf{0.781} \\
S1 RandomSplit & 0.497 & 0.723 & 0.499 & 0.821 & 0.916 & 0.748 \\
S2 NoGate & 0.575 & 0.740 & 0.561 & 0.832 & 0.909 & 0.770 \\
S3 NoInteraction & 0.512 & 0.729 & \textbf{0.574} & 0.809 & 0.910 & 0.758 \\
\hspace{3mm} S3.1 NoAlign & 0.562 & 0.732 & 0.554 & 0.839 & 0.912 & 0.768 \\
\hspace{3mm} S3.2 NoRefine & 0.472 & 0.709 & 0.571 & 0.827 & 0.920 & 0.752 \\
S4 SimpleFusion & 0.518 & 0.693 & 0.418 & 0.812 & 0.728 & 0.650 \\
S5 NoSelfTraining & 0.535 & 0.720 & 0.551 & 0.866 & 0.921 & 0.763 \\
\hspace{3mm} S5.1 NoKD & 0.492 & 0.723 & 0.560 & 0.797 & \textbf{0.944} & 0.765 \\
\hspace{3mm} S5.2 NoConn & 0.504 & 0.717 & 0.555 & 0.773 & 0.943 & 0.759 \\
S6 LinearEncoder & 0.508 & 0.719 & 0.551 & 0.771 & 0.942 & 0.759 \\
\bottomrule
\end{tabular}
}%
\end{table}


\begin{thebibliography}{42}


\ifx \showCODEN    \undefined \def \showCODEN     #1{\unskip}     \fi
\ifx \showISBNx    \undefined \def \showISBNx     #1{\unskip}     \fi
\ifx \showISBNxiii \undefined \def \showISBNxiii  #1{\unskip}     \fi
\ifx \showISSN     \undefined \def \showISSN      #1{\unskip}     \fi
\ifx \showLCCN     \undefined \def \showLCCN      #1{\unskip}     \fi
\ifx \shownote     \undefined \def \shownote      #1{#1}          \fi
\ifx \showarticletitle \undefined \def \showarticletitle #1{#1}   \fi
\ifx \showURL      \undefined \def \showURL       {\relax}        \fi
\providecommand\bibfield[2]{#2}
\providecommand\bibinfo[2]{#2}
\providecommand\natexlab[1]{#1}
\providecommand\showeprint[2][]{arXiv:#2}

\bibitem[Baek et~al\mbox{.}(2024)]%
        {baek2024cradle}
\bibfield{author}{\bibinfo{person}{Seungheun Baek}, \bibinfo{person}{Soyon
  Park}, \bibinfo{person}{Yan~Ting Chok}, \bibinfo{person}{Junhyun Lee},
  \bibinfo{person}{Jueon Park}, \bibinfo{person}{Mogan Gim}, {and}
  \bibinfo{person}{Jaewoo Kang}.} \bibinfo{year}{2024}\natexlab{}.
\newblock \bibinfo{title}{{CRADLE-VAE}: Enhancing Single-Cell Gene Perturbation
  Modeling with Counterfactual Reasoning-based Artifact Disentanglement}.
\newblock
\showeprint[arxiv]{2409.05484}~[cs.LG]
\href{https://doi.org/10.48550/arXiv.2409.05484}{doi:\nolinkurl{10.48550/arXiv.2409.05484}}
\newblock
\shownote{arXiv v2 (last revised 2024-09-10)}.


\bibitem[Baek et~al\mbox{.}(2025)]%
        {scfoundation_survey2024}
\bibfield{author}{\bibinfo{person}{Seungbyn Baek}, \bibinfo{person}{Kyungwoo
  Song}, {and} \bibinfo{person}{Insuk Lee}.} \bibinfo{year}{2025}\natexlab{}.
\newblock \showarticletitle{Single-cell foundation models: bringing artificial
  intelligence into cell biology}.
\newblock \bibinfo{journal}{\emph{Experimental \& Molecular Medicine}}
  \bibinfo{volume}{57} (\bibinfo{year}{2025}), \bibinfo{pages}{2169--2181}.
\newblock
\href{https://doi.org/10.1038/s12276-025-01547-5}{doi:\nolinkurl{10.1038/s12276-025-01547-5}}


\bibitem[Chen and He(2021a)]%
        {chen2021exploring}
\bibfield{author}{\bibinfo{person}{Xinlei Chen} {and} \bibinfo{person}{Kaiming
  He}.} \bibinfo{year}{2021}\natexlab{a}.
\newblock \showarticletitle{Exploring Simple Siamese Representation Learning}.
  In \bibinfo{booktitle}{\emph{Proceedings of the IEEE/CVF Conference on
  Computer Vision and Pattern Recognition (CVPR)}}. \bibinfo{publisher}{IEEE},
  \bibinfo{address}{Piscataway, NJ, USA}, \bibinfo{numpages}{11}~pages.
\newblock


\bibitem[Chen and He(2021b)]%
        {chen2021simsiam}
\bibfield{author}{\bibinfo{person}{Xinlei Chen} {and} \bibinfo{person}{Kaiming
  He}.} \bibinfo{year}{2021}\natexlab{b}.
\newblock \showarticletitle{Exploring Simple Siamese Representation Learning}.
  In \bibinfo{booktitle}{\emph{Proceedings of the IEEE/CVF Conference on
  Computer Vision and Pattern Recognition (CVPR)}}. \bibinfo{publisher}{IEEE},
  \bibinfo{address}{Piscataway, NJ, USA}, \bibinfo{numpages}{11}~pages.
\newblock


\bibitem[Chockalingam et~al\mbox{.}(2025)]%
        {chockalingam2025scement}
\bibfield{author}{\bibinfo{person}{Sriram~P. Chockalingam},
  \bibinfo{person}{Maneesha Aluru}, {and} \bibinfo{person}{Srinivas Aluru}.}
  \bibinfo{year}{2025}\natexlab{}.
\newblock \showarticletitle{SCEMENT: scalable and memory efficient integration
  of large-scale single-cell {RNA}-sequencing data}.
\newblock \bibinfo{journal}{\emph{Bioinformatics}} \bibinfo{volume}{41},
  \bibinfo{number}{2} (\bibinfo{year}{2025}), \bibinfo{pages}{btaf057}.
\newblock
\href{https://doi.org/10.1093/bioinformatics/btaf057}{doi:\nolinkurl{10.1093/bioinformatics/btaf057}}


\bibitem[Cui et~al\mbox{.}(2024)]%
        {scgpt2024}
\bibfield{author}{\bibinfo{person}{Haotian Cui}, \bibinfo{person}{Cheng Wang},
  \bibinfo{person}{Hammad Maan}, \bibinfo{person}{Kevin Pang},
  \bibinfo{person}{Feng Luo}, \bibinfo{person}{Nan Duan}, {and}
  \bibinfo{person}{Bo Wang}.} \bibinfo{year}{2024}\natexlab{}.
\newblock \showarticletitle{scGPT: toward building a foundation model for
  single-cell multi-omics using generative {AI}}.
\newblock \bibinfo{journal}{\emph{Nature Methods}} \bibinfo{volume}{21},
  \bibinfo{number}{8} (\bibinfo{year}{2024}), \bibinfo{pages}{1470--1480}.
\newblock
\href{https://doi.org/10.1038/s41592-024-02201-0}{doi:\nolinkurl{10.1038/s41592-024-02201-0}}


\bibitem[Gong et~al\mbox{.}(2016)]%
        {gong2016domain}
\bibfield{author}{\bibinfo{person}{Mingming Gong}, \bibinfo{person}{Kun Zhang},
  \bibinfo{person}{Tongliang Liu}, \bibinfo{person}{Dacheng Tao},
  \bibinfo{person}{Clark Glymour}, {and} \bibinfo{person}{Bernhard
  Sch{\"o}lkopf}.} \bibinfo{year}{2016}\natexlab{}.
\newblock \showarticletitle{Domain Adaptation with Conditional Transferable
  Components}. In \bibinfo{booktitle}{\emph{Proceedings of The 33rd
  International Conference on Machine Learning}}
  \emph{(\bibinfo{series}{Proceedings of Machine Learning Research},
  Vol.~\bibinfo{volume}{48})}, \bibfield{editor}{\bibinfo{person}{Maria~Florina
  Balcan} {and} \bibinfo{person}{Kilian~Q. Weinberger}} (Eds.).
  \bibinfo{publisher}{PMLR}, \bibinfo{address}{New York, New York, USA},
  \bibinfo{pages}{2839--2848}.
\newblock
\urldef\tempurl%
\url{https://proceedings.mlr.press/v48/gong16.html}
\showURL{%
\tempurl}


\bibitem[Grill et~al\mbox{.}(2020a)]%
        {grill2020bootstrap}
\bibfield{author}{\bibinfo{person}{Jean-Bastien Grill},
  \bibinfo{person}{Florian Strub}, \bibinfo{person}{Florent Altch{\'e}},
  {et~al\mbox{.}}} \bibinfo{year}{2020}\natexlab{a}.
\newblock \showarticletitle{Bootstrap Your Own Latent: A New Approach to
  Self-Supervised Learning}. In \bibinfo{booktitle}{\emph{Advances in Neural
  Information Processing Systems (NeurIPS)}}. \bibinfo{publisher}{Curran
  Associates, Inc.}, \bibinfo{address}{Red Hook, NY, USA},
  \bibinfo{pages}{21271--21284}.
\newblock


\bibitem[Grill et~al\mbox{.}(2020b)]%
        {grill2020byol}
\bibfield{author}{\bibinfo{person}{Jean-Bastien Grill},
  \bibinfo{person}{Florian Strub}, \bibinfo{person}{Florent Altch{\'e}},
  {et~al\mbox{.}}} \bibinfo{year}{2020}\natexlab{b}.
\newblock \showarticletitle{Bootstrap Your Own Latent: A New Approach to
  Self-Supervised Learning}. In \bibinfo{booktitle}{\emph{Advances in Neural
  Information Processing Systems (NeurIPS)}}. \bibinfo{publisher}{Curran
  Associates, Inc.}, \bibinfo{address}{Red Hook, NY, USA},
  \bibinfo{pages}{21271--21284}.
\newblock


\bibitem[Ha et~al\mbox{.}(2017)]%
        {ha2016hypernetworks}
\bibfield{author}{\bibinfo{person}{David Ha}, \bibinfo{person}{Andrew Dai},
  {and} \bibinfo{person}{Quoc~V. Le}.} \bibinfo{year}{2017}\natexlab{}.
\newblock \showarticletitle{HyperNetworks}. In
  \bibinfo{booktitle}{\emph{International Conference on Learning
  Representations (ICLR)}}. \bibinfo{publisher}{OpenReview.net},
  \bibinfo{address}{Toulon, France}, \bibinfo{numpages}{18}~pages.
\newblock


\bibitem[Haghverdi et~al\mbox{.}(2018)]%
        {haghverdi2018batch}
\bibfield{author}{\bibinfo{person}{Laleh Haghverdi}, \bibinfo{person}{Aaron
  T.~L. Lun}, \bibinfo{person}{Michael~D. Morgan}, {and}
  \bibinfo{person}{John~C. Marioni}.} \bibinfo{year}{2018}\natexlab{}.
\newblock \showarticletitle{Batch effects in single-cell {RNA}-sequencing data
  are corrected by matching mutual nearest neighbors}.
\newblock \bibinfo{journal}{\emph{Nature Biotechnology}} \bibinfo{volume}{36},
  \bibinfo{number}{5} (\bibinfo{year}{2018}), \bibinfo{pages}{421--427}.
\newblock
\href{https://doi.org/10.1038/nbt.4091}{doi:\nolinkurl{10.1038/nbt.4091}}


\bibitem[Hinton et~al\mbox{.}(2015)]%
        {hinton2015distilling}
\bibfield{author}{\bibinfo{person}{Geoffrey Hinton}, \bibinfo{person}{Oriol
  Vinyals}, {and} \bibinfo{person}{Jeff Dean}.}
  \bibinfo{year}{2015}\natexlab{}.
\newblock \bibinfo{title}{Distilling the Knowledge in a Neural Network}.
\newblock \bibinfo{howpublished}{arXiv preprint arXiv:1503.02531}.
\newblock
\newblock
\shownote{Presented at NeurIPS 2015 Deep Learning Workshop}.


\bibitem[Hrovatin et~al\mbox{.}(2025)]%
        {sysvi2024}
\bibfield{author}{\bibinfo{person}{Karin Hrovatin}, \bibinfo{person}{Amir~Ali
  Moinfar}, \bibinfo{person}{Luke Zappia}, \bibinfo{person}{Shrey Parikh},
  \bibinfo{person}{Alejandro Tejada~Lapuerta}, \bibinfo{person}{Ben Lengerich},
  \bibinfo{person}{Manolis Kellis}, {and} \bibinfo{person}{Fabian~J. Theis}.}
  \bibinfo{year}{2025}\natexlab{}.
\newblock \showarticletitle{Integrating single-cell {RNA}-seq datasets with
  substantial batch effects}.
\newblock \bibinfo{journal}{\emph{BMC Genomics}} \bibinfo{volume}{26},
  \bibinfo{number}{1} (\bibinfo{year}{2025}), \bibinfo{pages}{974}.
\newblock
\href{https://doi.org/10.1186/s12864-025-12126-3}{doi:\nolinkurl{10.1186/s12864-025-12126-3}}


\bibitem[Klicpera et~al\mbox{.}(2019)]%
        {klicpera2019appnp}
\bibfield{author}{\bibinfo{person}{Johannes Klicpera},
  \bibinfo{person}{Aleksandar Bojchevski}, {and} \bibinfo{person}{Stephan
  G{\"u}nnemann}.} \bibinfo{year}{2019}\natexlab{}.
\newblock \showarticletitle{Predict then Propagate: Graph Neural Networks meet
  Personalized PageRank}. In \bibinfo{booktitle}{\emph{International Conference
  on Learning Representations (ICLR)}}. \bibinfo{publisher}{OpenReview.net},
  \bibinfo{address}{New Orleans, LA, USA}, \bibinfo{numpages}{15}~pages.
\newblock


\bibitem[Koenig et~al\mbox{.}(2022)]%
        {koenig2022single}
\bibfield{author}{\bibinfo{person}{Andrew~L Koenig}, \bibinfo{person}{Irina
  Shchukina}, \bibinfo{person}{Junedh Amrute}, \bibinfo{person}{Patrick~S
  Anber}, \bibinfo{person}{Katarina Yaber}, \bibinfo{person}{Kory~J Lavine},
  \bibinfo{person}{Sanjoy Bhattacharya}, \bibinfo{person}{Kamran~A Bhojani},
  \bibinfo{person}{Jesse Koenig}, {et~al\mbox{.}}}
  \bibinfo{year}{2022}\natexlab{}.
\newblock \showarticletitle{Single-cell transcriptomics reveals
  cell-type-specific diversification in human heart failure}.
\newblock \bibinfo{journal}{\emph{Nature Cardiovascular Research}}
  \bibinfo{volume}{1}, \bibinfo{number}{3} (\bibinfo{year}{2022}),
  \bibinfo{pages}{263--280}.
\newblock
\href{https://doi.org/10.1038/s44161-022-00028-6}{doi:\nolinkurl{10.1038/s44161-022-00028-6}}


\bibitem[Korsunsky et~al\mbox{.}(2019a)]%
        {korsunsky2019harmony}
\bibfield{author}{\bibinfo{person}{Ilya Korsunsky}, \bibinfo{person}{Nghia
  Millard}, \bibinfo{person}{Jean Fan}, \bibinfo{person}{Kamil Slowikowski},
  \bibinfo{person}{Fan Zhang}, \bibinfo{person}{Kevin Wei},
  \bibinfo{person}{Yuriy Baglaenko}, \bibinfo{person}{Michael Brenner},
  \bibinfo{person}{Po-Ru Loh}, {and} \bibinfo{person}{Soumya Raychaudhuri}.}
  \bibinfo{year}{2019}\natexlab{a}.
\newblock \showarticletitle{Fast, sensitive and accurate integration of
  single-cell data with {Harmony}}.
\newblock \bibinfo{journal}{\emph{Nature Methods}} \bibinfo{volume}{16},
  \bibinfo{number}{12} (\bibinfo{year}{2019}), \bibinfo{pages}{1289--1296}.
\newblock
\href{https://doi.org/10.1038/s41592-019-0619-0}{doi:\nolinkurl{10.1038/s41592-019-0619-0}}


\bibitem[Korsunsky et~al\mbox{.}(2019b)]%
        {korsunsky2019fast}
\bibfield{author}{\bibinfo{person}{Ilya Korsunsky}, \bibinfo{person}{Nghia
  Millard}, \bibinfo{person}{Jean Fan}, \bibinfo{person}{Kamil Slowikowski},
  \bibinfo{person}{Fan Zhang}, \bibinfo{person}{Kevin Wei},
  \bibinfo{person}{Yuriy Baglaenko}, \bibinfo{person}{Michael Brenner},
  \bibinfo{person}{Po-Ru Loh}, {and} \bibinfo{person}{Soumya Raychaudhuri}.}
  \bibinfo{year}{2019}\natexlab{b}.
\newblock \showarticletitle{Fast, sensitive and accurate integration of
  single-cell data with {Harmony}}.
\newblock \bibinfo{journal}{\emph{Nature Methods}} \bibinfo{volume}{16},
  \bibinfo{number}{12} (\bibinfo{year}{2019}), \bibinfo{pages}{1289--1296}.
\newblock
\href{https://doi.org/10.1038/s41592-019-0619-0}{doi:\nolinkurl{10.1038/s41592-019-0619-0}}


\bibitem[Leek et~al\mbox{.}(2010)]%
        {leek2010tackling}
\bibfield{author}{\bibinfo{person}{Jeffrey~T. Leek}, \bibinfo{person}{Robert~B.
  Scharpf}, \bibinfo{person}{H{\'e}ctor~Corrada Bravo}, \bibinfo{person}{David
  Simcha}, \bibinfo{person}{Benjamin Langmead}, \bibinfo{person}{W.~Evan
  Johnson}, \bibinfo{person}{Donald Geman}, \bibinfo{person}{Keith Baggerly},
  {and} \bibinfo{person}{Rafael~A. Irizarry}.} \bibinfo{year}{2010}\natexlab{}.
\newblock \showarticletitle{Tackling the widespread and critical impact of
  batch effects in high-throughput data}.
\newblock \bibinfo{journal}{\emph{Nature Reviews Genetics}}
  \bibinfo{volume}{11}, \bibinfo{number}{10} (\bibinfo{year}{2010}),
  \bibinfo{pages}{733--739}.
\newblock
\href{https://doi.org/10.1038/nrg2825}{doi:\nolinkurl{10.1038/nrg2825}}


\bibitem[Liu et~al\mbox{.}(2025)]%
        {liu2025biobatchnet}
\bibfield{author}{\bibinfo{person}{Haiping Liu}, \bibinfo{person}{Shaojie
  Zhang}, \bibinfo{person}{Shengzhong Mao}, \bibinfo{person}{Qian Zhao},
  \bibinfo{person}{Yuxi Zhou}, \bibinfo{person}{Andrew~P. Gilmore},
  \bibinfo{person}{Mauricio~A. Alvarez}, {and} \bibinfo{person}{Hongpeng
  Zhou}.} \bibinfo{year}{2025}\natexlab{}.
\newblock \bibinfo{title}{{BioBatchNet}: A Dual-Encoder Framework for Robust
  Batch Effect Correction in Imaging Mass Cytometry}.
\newblock \bibinfo{howpublished}{bioRxiv preprint}.
\newblock
\href{https://doi.org/10.1101/2025.03.15.643447}{doi:\nolinkurl{10.1101/2025.03.15.643447}}


\bibitem[Lopez et~al\mbox{.}(2018)]%
        {lopez2018deep}
\bibfield{author}{\bibinfo{person}{Romain Lopez}, \bibinfo{person}{Jeffrey
  Regier}, \bibinfo{person}{Michael~B. Cole}, \bibinfo{person}{Michael~I.
  Jordan}, {and} \bibinfo{person}{Nir Yosef}.} \bibinfo{year}{2018}\natexlab{}.
\newblock \showarticletitle{Deep generative modeling for single-cell
  transcriptomics}.
\newblock \bibinfo{journal}{\emph{Nature Methods}} \bibinfo{volume}{15},
  \bibinfo{number}{12} (\bibinfo{year}{2018}), \bibinfo{pages}{1053--1058}.
\newblock
\href{https://doi.org/10.1038/s41592-018-0229-2}{doi:\nolinkurl{10.1038/s41592-018-0229-2}}


\bibitem[Lotfollahi et~al\mbox{.}(2019)]%
        {lotfollahi2019scgen}
\bibfield{author}{\bibinfo{person}{Mohammad Lotfollahi},
  \bibinfo{person}{F.~Alexander Wolf}, {and} \bibinfo{person}{Fabian~J.
  Theis}.} \bibinfo{year}{2019}\natexlab{}.
\newblock \showarticletitle{{scGen} predicts single-cell perturbation
  responses}.
\newblock \bibinfo{journal}{\emph{Nature Methods}} \bibinfo{volume}{16},
  \bibinfo{number}{8} (\bibinfo{year}{2019}), \bibinfo{pages}{715--721}.
\newblock
\href{https://doi.org/10.1038/s41592-019-0494-8}{doi:\nolinkurl{10.1038/s41592-019-0494-8}}


\bibitem[Luecken et~al\mbox{.}(2022)]%
        {luecken2022benchmarking}
\bibfield{author}{\bibinfo{person}{Malte~D. Luecken}, \bibinfo{person}{M.
  B{\"u}ttner}, \bibinfo{person}{K. Chaichoompu}, \bibinfo{person}{A. Danese},
  \bibinfo{person}{M. Interlandi}, \bibinfo{person}{M.~F. Mueller},
  \bibinfo{person}{D.~C. Strobl}, \bibinfo{person}{L. Zappia},
  \bibinfo{person}{M. Dugas}, \bibinfo{person}{M. Colom{\'e}-Tatch{\'e}}, {and}
  \bibinfo{person}{Fabian~J. Theis}.} \bibinfo{year}{2022}\natexlab{}.
\newblock \showarticletitle{Benchmarking atlas-level data integration in
  single-cell genomics}.
\newblock \bibinfo{journal}{\emph{Nature Methods}} \bibinfo{volume}{19},
  \bibinfo{number}{1} (\bibinfo{year}{2022}), \bibinfo{pages}{41--50}.
\newblock
\href{https://doi.org/10.1038/s41592-021-01336-8}{doi:\nolinkurl{10.1038/s41592-021-01336-8}}


\bibitem[Makino et~al\mbox{.}(2025)]%
        {makino2025supervised}
\bibfield{author}{\bibinfo{person}{Taro Makino}, \bibinfo{person}{Ji~Won Park},
  \bibinfo{person}{Natasa Tagasovska}, \bibinfo{person}{Takamasa Kudo},
  \bibinfo{person}{Paula Coelho}, \bibinfo{person}{Jan-Christian Huetter},
  \bibinfo{person}{Heming Yao}, \bibinfo{person}{Burkhard Hoeckendorf},
  \bibinfo{person}{Ana~Carolina Leote}, \bibinfo{person}{Stephen Ra},
  \bibinfo{person}{David Richmond}, \bibinfo{person}{Kyunghyun Cho},
  \bibinfo{person}{Aviv Regev}, {and} \bibinfo{person}{Romain Lopez}.}
  \bibinfo{year}{2025}\natexlab{}.
\newblock \bibinfo{title}{Supervised Contrastive Block Disentanglement}.
\newblock
\showeprint[arxiv]{2502.07281}~[cs.LG]
\href{https://doi.org/10.48550/arXiv.2502.07281}{doi:\nolinkurl{10.48550/arXiv.2502.07281}}
\newblock
\shownote{arXiv v1 (submitted 2025-02-11)}.


\bibitem[Mao et~al\mbox{.}(2024)]%
        {mao2024learning}
\bibfield{author}{\bibinfo{person}{Haiyi Mao}, \bibinfo{person}{Romain Lopez},
  \bibinfo{person}{Kai Liu}, \bibinfo{person}{Jan-Christian Huetter},
  \bibinfo{person}{David Richmond}, \bibinfo{person}{Panayiotis~V. Benos},
  {and} \bibinfo{person}{Lin Qiu}.} \bibinfo{year}{2024}\natexlab{}.
\newblock \bibinfo{title}{Learning Identifiable Factorized Causal
  Representations of Cellular Responses}.
\newblock
\showeprint[arxiv]{2410.22472}~[cs.LG]
\href{https://doi.org/10.48550/arXiv.2410.22472}{doi:\nolinkurl{10.48550/arXiv.2410.22472}}
\newblock
\shownote{arXiv v3 (last revised 2024-12-02)}.


\bibitem[Muraro et~al\mbox{.}(2016)]%
        {muraro2016single}
\bibfield{author}{\bibinfo{person}{Mauro~J Muraro}, \bibinfo{person}{Gitanjali
  Dharmadhikari}, \bibinfo{person}{Dominic Gr{\"u}n}, \bibinfo{person}{Nathalie
  Groen}, \bibinfo{person}{Tim Dielen}, \bibinfo{person}{Erik Jansen},
  \bibinfo{person}{Leon Van~Gurp}, \bibinfo{person}{Marten~A Engelse},
  \bibinfo{person}{Francoise Carlotti}, \bibinfo{person}{Eelco~Jp De~Koning},
  {et~al\mbox{.}}} \bibinfo{year}{2016}\natexlab{}.
\newblock \showarticletitle{A single-cell transcriptome atlas of the human
  pancreas}.
\newblock \bibinfo{journal}{\emph{Cell systems}} \bibinfo{volume}{3},
  \bibinfo{number}{4} (\bibinfo{year}{2016}), \bibinfo{pages}{385--394}.
\newblock


\bibitem[Piran et~al\mbox{.}(2024)]%
        {piran2024disentanglement}
\bibfield{author}{\bibinfo{person}{Zohar Piran}, \bibinfo{person}{Noga Cohen},
  \bibinfo{person}{Yedid Hoshen}, {and} \bibinfo{person}{Mor Nitzan}.}
  \bibinfo{year}{2024}\natexlab{}.
\newblock \showarticletitle{Disentanglement of single-cell data with
  {biolord}}.
\newblock \bibinfo{journal}{\emph{Nature Biotechnology}}  \bibinfo{volume}{42}
  (\bibinfo{year}{2024}), \bibinfo{pages}{766--774}.
\newblock
\href{https://doi.org/10.1038/s41587-023-01927-2}{doi:\nolinkurl{10.1038/s41587-023-01927-2}}


\bibitem[Simon et~al\mbox{.}(2021)]%
        {simon2021insct}
\bibfield{author}{\bibinfo{person}{Lukas~M. Simon}, \bibinfo{person}{Yin-Ying
  Wang}, {and} \bibinfo{person}{Zhongming Zhao}.}
  \bibinfo{year}{2021}\natexlab{}.
\newblock \showarticletitle{Integration of millions of transcriptomes using
  batch-aware triplet neural networks}.
\newblock \bibinfo{journal}{\emph{Nature Machine Intelligence}}
  \bibinfo{volume}{3} (\bibinfo{year}{2021}), \bibinfo{pages}{705--715}.
\newblock
\href{https://doi.org/10.1038/s42256-021-00361-8}{doi:\nolinkurl{10.1038/s42256-021-00361-8}}


\bibitem[Stuart et~al\mbox{.}(2019)]%
        {stuart2019comprehensive}
\bibfield{author}{\bibinfo{person}{Tim Stuart}, \bibinfo{person}{Andrew
  Butler}, \bibinfo{person}{Paul Hoffman}, \bibinfo{person}{Christoph
  Hafemeister}, \bibinfo{person}{Efthymia Papalexi},
  \bibinfo{person}{William~M. Mauck}, \bibinfo{person}{Yuhan Hao},
  \bibinfo{person}{Marlon Stoeckius}, \bibinfo{person}{Peter Smibert}, {and}
  \bibinfo{person}{Rahul Satija}.} \bibinfo{year}{2019}\natexlab{}.
\newblock \showarticletitle{Comprehensive integration of single-cell data}.
\newblock \bibinfo{journal}{\emph{Cell}} \bibinfo{volume}{177},
  \bibinfo{number}{7} (\bibinfo{year}{2019}), \bibinfo{pages}{1888--1902.e21}.
\newblock
\href{https://doi.org/10.1016/j.cell.2019.05.031}{doi:\nolinkurl{10.1016/j.cell.2019.05.031}}


\bibitem[Tarvainen and Valpola(2017)]%
        {tarvainen2017mean}
\bibfield{author}{\bibinfo{person}{Antti Tarvainen} {and}
  \bibinfo{person}{Harri Valpola}.} \bibinfo{year}{2017}\natexlab{}.
\newblock \showarticletitle{Mean teachers are better role models:
  Weight-averaged consistency targets improve semi-supervised deep learning
  results}. In \bibinfo{booktitle}{\emph{Advances in Neural Information
  Processing Systems (NeurIPS)}}. \bibinfo{publisher}{Curran Associates, Inc.},
  \bibinfo{address}{Red Hook, NY, USA}, \bibinfo{pages}{1195--1204}.
\newblock


\bibitem[{The Tabula Muris Consortium}(2020)]%
        {tabula2020single}
\bibfield{author}{\bibinfo{person}{{The Tabula Muris Consortium}}.}
  \bibinfo{year}{2020}\natexlab{}.
\newblock \showarticletitle{A single-cell transcriptomic atlas characterizes
  ageing tissues in the mouse}.
\newblock \bibinfo{journal}{\emph{Nature}} \bibinfo{volume}{583},
  \bibinfo{number}{7817} (\bibinfo{year}{2020}), \bibinfo{pages}{590--595}.
\newblock
\href{https://doi.org/10.1038/s41586-020-2496-1}{doi:\nolinkurl{10.1038/s41586-020-2496-1}}


\bibitem[Theodoris et~al\mbox{.}(2023)]%
        {theodoris2023geneformer}
\bibfield{author}{\bibinfo{person}{Christina~V. Theodoris},
  \bibinfo{person}{Ling Xiao}, \bibinfo{person}{Akash Chopra}, {et~al\mbox{.}}}
  \bibinfo{year}{2023}\natexlab{}.
\newblock \showarticletitle{Transfer learning enables predictions in network
  biology}.
\newblock \bibinfo{journal}{\emph{Nature}} \bibinfo{volume}{618},
  \bibinfo{number}{7965} (\bibinfo{year}{2023}), \bibinfo{pages}{616--624}.
\newblock
\href{https://doi.org/10.1038/s41586-023-06139-9}{doi:\nolinkurl{10.1038/s41586-023-06139-9}}


\bibitem[Traag et~al\mbox{.}(2019)]%
        {traag2019from}
\bibfield{author}{\bibinfo{person}{Vincent~A. Traag}, \bibinfo{person}{Ludo
  Waltman}, {and} \bibinfo{person}{Nees~Jan Van~Eck}.}
  \bibinfo{year}{2019}\natexlab{}.
\newblock \showarticletitle{From Louvain to Leiden: guaranteeing well-connected
  communities}.
\newblock \bibinfo{journal}{\emph{Scientific Reports}} \bibinfo{volume}{9},
  \bibinfo{number}{1} (\bibinfo{year}{2019}), \bibinfo{pages}{5233}.
\newblock
\href{https://doi.org/10.1038/s41598-019-41695-z}{doi:\nolinkurl{10.1038/s41598-019-41695-z}}


\bibitem[Trapnell(2015)]%
        {trapnell2015defining}
\bibfield{author}{\bibinfo{person}{Cole Trapnell}.}
  \bibinfo{year}{2015}\natexlab{}.
\newblock \showarticletitle{Defining cell types and states with single-cell
  genomics}.
\newblock \bibinfo{journal}{\emph{Genome Research}} \bibinfo{volume}{25},
  \bibinfo{number}{10} (\bibinfo{year}{2015}), \bibinfo{pages}{1491--1498}.
\newblock
\href{https://doi.org/10.1101/gr.190595.115}{doi:\nolinkurl{10.1101/gr.190595.115}}


\bibitem[Travaglini et~al\mbox{.}(2020)]%
        {travaglini2020molecular}
\bibfield{author}{\bibinfo{person}{Kyle~J Travaglini}, \bibinfo{person}{Ahmad~N
  Nabhan}, \bibinfo{person}{Lolita Penland}, \bibinfo{person}{Rahul Sinha},
  \bibinfo{person}{Astrid Gillich}, \bibinfo{person}{Rene~V Sit},
  \bibinfo{person}{Stephen Chang}, \bibinfo{person}{Stephanie~D Conley},
  \bibinfo{person}{Yelena Mber}, \bibinfo{person}{Mia Huff}, {et~al\mbox{.}}}
  \bibinfo{year}{2020}\natexlab{}.
\newblock \showarticletitle{A molecular cell atlas of the human lung from
  single-cell RNA sequencing}.
\newblock \bibinfo{journal}{\emph{Nature}} \bibinfo{volume}{587},
  \bibinfo{number}{7835} (\bibinfo{year}{2020}), \bibinfo{pages}{619--625}.
\newblock
\href{https://doi.org/10.1038/s41586-020-2922-4}{doi:\nolinkurl{10.1038/s41586-020-2922-4}}


\bibitem[Vaswani et~al\mbox{.}(2017)]%
        {vaswani2017attention}
\bibfield{author}{\bibinfo{person}{Ashish Vaswani}, \bibinfo{person}{Noam
  Shazeer}, \bibinfo{person}{Niki Parmar}, \bibinfo{person}{Jakob Uszkoreit},
  \bibinfo{person}{Llion Jones}, \bibinfo{person}{Aidan~N. Gomez},
  \bibinfo{person}{{\L}ukasz Kaiser}, {and} \bibinfo{person}{Illia
  Polosukhin}.} \bibinfo{year}{2017}\natexlab{}.
\newblock \showarticletitle{Attention is All you Need}. In
  \bibinfo{booktitle}{\emph{Advances in Neural Information Processing Systems
  (NeurIPS)}}. \bibinfo{publisher}{Curran Associates, Inc.},
  \bibinfo{address}{Red Hook, NY, USA}, \bibinfo{pages}{5998--6008}.
\newblock


\bibitem[Wang et~al\mbox{.}(2021)]%
        {wang2021scgnn}
\bibfield{author}{\bibinfo{person}{Juexin Wang}, \bibinfo{person}{Anjun Ma},
  \bibinfo{person}{Yuzhou Chang}, \bibinfo{person}{Jianting Gong},
  \bibinfo{person}{Yuexu Jiang}, \bibinfo{person}{Ren Qi},
  \bibinfo{person}{Cankun Wang}, \bibinfo{person}{Hongjun Fu},
  \bibinfo{person}{Qin Ma}, {and} \bibinfo{person}{Dong Xu}.}
  \bibinfo{year}{2021}\natexlab{}.
\newblock \showarticletitle{{scGNN} is a novel graph neural network framework
  for single-cell {RNA}-Seq analyses}.
\newblock \bibinfo{journal}{\emph{Nature Communications}} \bibinfo{volume}{12},
  \bibinfo{number}{1} (\bibinfo{year}{2021}), \bibinfo{pages}{1882}.
\newblock
\href{https://doi.org/10.1038/s41467-021-22197-x}{doi:\nolinkurl{10.1038/s41467-021-22197-x}}


\bibitem[Wolf et~al\mbox{.}(2018)]%
        {wolf2018scanpy}
\bibfield{author}{\bibinfo{person}{F.~Alexander Wolf}, \bibinfo{person}{Philipp
  Angerer}, {and} \bibinfo{person}{Fabian~J. Theis}.}
  \bibinfo{year}{2018}\natexlab{}.
\newblock \showarticletitle{{SCANPY}: large-scale single-cell gene expression
  data analysis}.
\newblock \bibinfo{journal}{\emph{Genome Biology}} \bibinfo{volume}{19},
  \bibinfo{number}{1} (\bibinfo{year}{2018}), \bibinfo{pages}{15}.
\newblock
\href{https://doi.org/10.1186/s13059-017-1382-0}{doi:\nolinkurl{10.1186/s13059-017-1382-0}}


\bibitem[Xiong et~al\mbox{.}(2022)]%
        {xiong2022scalex}
\bibfield{author}{\bibinfo{person}{Lei Xiong}, \bibinfo{person}{Kang Tian},
  \bibinfo{person}{Yuzhe Li}, \bibinfo{person}{Weixi Ning},
  \bibinfo{person}{Xin Gao}, {and} \bibinfo{person}{Qiangfeng~Cliff Zhang}.}
  \bibinfo{year}{2022}\natexlab{}.
\newblock \showarticletitle{Online single-cell data integration through
  projecting heterogeneous datasets into a common cell-embedding space}.
\newblock \bibinfo{journal}{\emph{Nature Communications}}  \bibinfo{volume}{13}
  (\bibinfo{year}{2022}), \bibinfo{pages}{6118}.
\newblock
\href{https://doi.org/10.1038/s41467-022-33758-z}{doi:\nolinkurl{10.1038/s41467-022-33758-z}}


\bibitem[Xu et~al\mbox{.}(2021)]%
        {xu2021probabilistic}
\bibfield{author}{\bibinfo{person}{Chenling Xu}, \bibinfo{person}{Romain
  Lopez}, \bibinfo{person}{Edouard Mehlman}, \bibinfo{person}{Jeffrey Regier},
  \bibinfo{person}{Michael~I. Jordan}, {and} \bibinfo{person}{Nir Yosef}.}
  \bibinfo{year}{2021}\natexlab{}.
\newblock \showarticletitle{Probabilistic harmonization and annotation of
  single-cell transcriptomics data with deep generative models}.
\newblock \bibinfo{journal}{\emph{Molecular Systems Biology}}
  \bibinfo{volume}{17}, \bibinfo{number}{1} (\bibinfo{year}{2021}),
  \bibinfo{pages}{e9620}.
\newblock
\href{https://doi.org/10.15252/msb.20209620}{doi:\nolinkurl{10.15252/msb.20209620}}


\bibitem[Zhao et~al\mbox{.}(2025)]%
        {sccobra2025}
\bibfield{author}{\bibinfo{person}{Bowen Zhao}, \bibinfo{person}{Kailu Song},
  \bibinfo{person}{Dong-Qing Wei}, \bibinfo{person}{Yi Xiong}, {and}
  \bibinfo{person}{Jun Ding}.} \bibinfo{year}{2025}\natexlab{}.
\newblock \showarticletitle{{scCobra} allows contrastive cell embedding
  learning with domain adaptation for single cell data integration and
  harmonization}.
\newblock \bibinfo{journal}{\emph{Communications Biology}}  \bibinfo{volume}{8}
  (\bibinfo{year}{2025}), \bibinfo{pages}{233}.
\newblock
\href{https://doi.org/10.1038/s42003-025-07692-x}{doi:\nolinkurl{10.1038/s42003-025-07692-x}}


\bibitem[Zhou et~al\mbox{.}(2025a)]%
        {zhou2025gte}
\bibfield{author}{\bibinfo{person}{Yang Zhou}, \bibinfo{person}{Qiongyu Sheng},
  \bibinfo{person}{Guohua Wang}, \bibinfo{person}{Li Xu}, {and}
  \bibinfo{person}{Shuilin Jin}.} \bibinfo{year}{2025}\natexlab{a}.
\newblock \showarticletitle{Quantifying batch effects for individual genes in
  single-cell data}.
\newblock \bibinfo{journal}{\emph{Nature Computational Science}}
  \bibinfo{volume}{5}, \bibinfo{number}{8} (\bibinfo{year}{2025}),
  \bibinfo{pages}{612--620}.
\newblock
\href{https://doi.org/10.1038/s43588-025-00824-7}{doi:\nolinkurl{10.1038/s43588-025-00824-7}}


\bibitem[Zhou et~al\mbox{.}(2025b)]%
        {zhou2025quantifying}
\bibfield{author}{\bibinfo{person}{Yang Zhou}, \bibinfo{person}{Qiongyu Sheng},
  \bibinfo{person}{Guohua Wang}, \bibinfo{person}{Li Xu}, {and}
  \bibinfo{person}{Shuilin Jin}.} \bibinfo{year}{2025}\natexlab{b}.
\newblock \showarticletitle{Quantifying batch effects for individual genes in
  single-cell data}.
\newblock \bibinfo{journal}{\emph{Nature Computational Science}}
  \bibinfo{volume}{5}, \bibinfo{number}{8} (\bibinfo{year}{2025}),
  \bibinfo{pages}{612--620}.
\newblock
\href{https://doi.org/10.1038/s43588-025-00824-7}{doi:\nolinkurl{10.1038/s43588-025-00824-7}}


\end{thebibliography}
\end{document}